\documentclass[sigconf]{acmart}
\AtBeginDocument{
  \providecommand\BibTeX{{
    \normalfont B\kern-0.5em{\scshape i\kern-0.25em b}\kern-0.8em\TeX}}}

\newcommand{\smallsec}[1]{\vspace{0.05in} \noindent {\bf #1.}}

\copyrightyear{2020}
\acmYear{2020}
\setcopyright{acmlicensed}
\acmConference[FAT* '20]{Conference on Fairness, Accountability, and Transparency}{January 27--30, 2020}{Barcelona, Spain}
\acmBooktitle{Conference on Fairness, Accountability, and Transparency (FAT* '20), January 27--30, 2020, Barcelona, Spain}
\acmPrice{15.00}
\acmDOI{10.1145/3351095.3375709}
\acmISBN{978-1-4503-6936-7/20/02}
\begin{document}

\title{Towards Fairer Datasets: Filtering and Balancing the Distribution of the People Subtree in the ImageNet Hierarchy}

\author{Kaiyu Yang}
\orcid{0000-0002-2777-612X}
\affiliation{
  \institution{Princeton University}
  \streetaddress{35 Olden Street}
  \city{Princeton, NJ}
  }
\email{kaiyuy@cs.princeton.edu}

\author{Klint Qinami}
\affiliation{
  \institution{Princeton University}
  \streetaddress{35 Olden Street}
  \city{Princeton, NJ}
  }
\email{kqinami@cs.princeton.edu}

\author{Li Fei-Fei}
\affiliation{
  \institution{Stanford University}
  \streetaddress{353 Serra Mall}
  \city{Stanford, CA}
  }
\email{feifeili@cs.stanford.edu}

\author{Jia Deng}
\affiliation{
  \institution{Princeton University}
  \streetaddress{35 Olden Street}
  \city{Princeton, NJ}
  }
\email{jiadeng@cs.princeton.edu}

\author{Olga Russakovsky}
\affiliation{
  \institution{Princeton University}
  \streetaddress{35 Olden Street}
  \city{Princeton, NJ}
  }
\email{olgarus@cs.princeton.edu}

\begin{abstract}
Computer vision technology is being used by many but remains representative of only a few. People have reported misbehavior of computer vision models, including offensive prediction results and lower performance for underrepresented groups. Current computer vision models are typically developed using datasets consisting of manually annotated images or videos; the data and label distributions in these datasets are critical to the models' behavior. In this paper, we examine ImageNet, a large-scale ontology of images that has spurred the development of many modern computer vision methods. We consider three key factors within the \texttt{person} subtree of ImageNet that may lead to problematic behavior in downstream computer vision technology: (1) the stagnant concept vocabulary of WordNet, (2) the attempt at exhaustive illustration of all categories with images, and (3) the inequality of representation in the images within concepts. We seek to illuminate the root causes of these concerns and take the first steps to mitigate them constructively.
\end{abstract}

\maketitle

\section{Introduction}
As computer vision technology becomes widespread in people's Internet experience and daily lives, it is increasingly important for computer vision models to produce results that are appropriate and fair. However, there are notorious and persistent issues. For example, face recognition systems have been demonstrated to have disproportionate error rates across race groups, in part attributed to the underrepresentation of some skin tones in face recognition datasets~\cite{buolamwini2018gender}. Models for recognizing human activities perpetuate gender biases after seeing the strong correlations between gender and activity in the data~\cite{zhao2017men,hendricks2018women}. The downstream effects range from perpetuating harmful stereotypes~\cite{noble2018algorithms} to increasing the likelihood of being unfairly suspected of a crime (e.g., when face recognition models are used in surveillance cameras). 

Many of these concerns can be traced back to the datasets used to train the computer vision models. Thus, questions of fairness and representation in datasets have come to the forefront. In this work, we focus on one dataset, ImageNet~\cite{deng2009imagenet}, which has arguably been the most influential dataset of the modern era of deep learning in computer vision. ImageNet is a large-scale image ontology collected to enable the development of robust visual recognition models. The dataset spearheaded multiple breakthroughs in object recognition~\cite{krizhevsky2012imagenet, simonyan2014very, he2016deep}. In addition, the feature representation learned on ImageNet images has been used as a backbone for a variety of computer vision tasks such as object detection~\cite{girshick2015fast, redmon2016you}, human activity understanding~\cite{simonyan2014two}, image captioning~\cite{vinyals2015show}, and recovering depth from a single RGB image~\cite{liu2015deep}, to name a few. Works such as Huh et al.~\cite{huh2016makes} have analyzed the factors that contributed to ImageNet's wide adoption. Despite remaining a free education dataset released for non-commercial use only,\footnote{Please refer to ImageNet terms and conditions at \url{image-net.org/download-faq}} the dataset has had profound impact on both academic and industrial research.

With ImageNet's large scale and diverse use cases, we examine the potential social concerns or biases that may be reflected or amplified in its data. It is important to note here that references to ``ImageNet'' typically imply a subset of 1,000 categories selected for the image classification task in the ImageNet Large Scale Visual Recognition Challenge (ILSVRC) of 2012-2017~\cite{russakovsky2015imagenet}, and much of the research has focused on this subset of the data. So far, Dulhanty and Wong~\cite{dulhanty2019auditing} studied the demographics of people in ILSVRC data by using computer vision models to predict the gender and age of depicted people, and demonstrated that, e.g., males aged 15 to 29 make up the largest subgroup.\footnote{As noted by the authors themselves, this approach poses a chicken-or-egg problem as trained models are likely to exhibit gender or age biases, thus limiting their ability to accurately benchmark dataset bias.} Stock and Cisse~\cite{stock2018convnets} did not explicitly analyze the dataset but demonstrate that models trained on ILSVRC exhibit misclassifications consistent with racial stereotypes. Shankar et al.~\cite{shankar2017no} and DeVries et al.~\cite{devries2019does} showed that most images come from Europe and the United States, and the resulting models have difficulty generalizing to images from other places. Overall, these studies identify a small handful of protected attributes and analyze their distribution and/or impact within the ILSVRC dataset, with the goal of \emph{illuminating} the existing bias.

\smallsec{Goals and contributions} There are two key distinctions of our work. First, we look beyond ILSVRC~\cite{russakovsky2015imagenet} to the broader ImageNet dataset~\cite{deng2009imagenet}. As model accuracy on the challenge benchmark is now near-perfect, it is time to examine the larger setting of ImageNet. The 1,000 categories selected for the challenge contain only 3 people categories (\texttt{scuba diver}, \texttt{bridegroom}, and \texttt{baseball player}) while the full ImageNet contains 2,832 people categories under the \texttt{person} subtree (accounting for roughly 8.3\% of the total images). Their use can be problematic and raises important questions about fairness and representation. In this work, we focus on the \texttt{person} subtree of the full ImageNet hierarchy.

Second, in contrast to prior work, our goal is to do a deeper analysis of the root causes of bias and misrepresentation, and to propose concrete steps towards mitigating them. We identify three key factors that may lead to problematic behavior in downstream technology: (1) the stagnant concept vocabulary from WordNet~\cite{miller1998wordnet}, (2) the attempt at exhaustive illustration of all categories with images, and (3) the inequality of demographic representation in the images. For each factor, we seek to illuminate the root causes of the concern, take the first steps to mitigate them through carefully designed annotation procedures, and identify future paths towards comprehensively addressing the underlying issues.

Concretely, we thoroughly analyze the \texttt{person} subtree of ImageNet and plan to modify it along several dimensions. First, in Sec.~\ref{sec:synsets}, we examine the 2,832 people categories that are annotated within the subtree, and determine that 1,593 of them are potentially offensive labels that should not be used in the context of an image recognition dataset. We plan to remove all of these from ImageNet. Second, in Sec.~\ref{sec:imageability}, out of the remaining 1,239 categories we find that only 158 of them are visual, with the remaining categories simply demonstrating annotators' bias. We recommend further filtering the \texttt{person} subtree down to only these 158 categories when training visual recognition models. Finally, in Sec.~\ref{sec:attributes} we run a large-scale crowdsourcing study to manually annotate the gender, skin color, and age of the people depicted in ImageNet images corresponding to these remaining categories. While the individual annotations may be imperfect despite our best efforts (e.g., the annotated gender expression may not correspond to the depicted person's gender identity), we can nevertheless compute the approximate demographic breakdown. We believe that releasing these sensitive attribute annotations directly is not the right step for ethical reasons, and instead plan to release a Web interface that allows an interested user to \emph{filter} the images within a category to achieve a new target demographic distribution.

We are working on incorporating these suggestions into the dataset. We will additionally release our annotation interfaces\footnote{\url{image-net.org/filtering-and-balancing}} to allow for similar cleanup of other computer vision benchmarks.

\section{Related Work on Fairness in Machine Learning}

We begin with a more general look at the literature that identified or attempted to mitigate bias in modern artificial intelligence systems. In short, datasets often have biased distributions of demographics (gender, race, age, etc.); machine learning models are trained to exploit whatever correlations exist in the data, leading to discriminatory behavior against underrepresented groups~\cite{barocas2016big, barocas2017big}. A great overview of the history of machine learning fairness can be found in Hutchinson and Mitchell~\cite{hutchinson2019history}. The approaches to address fairness concerns fall loosely into two categories: (1) identifying and correcting issues in datasets or (2) studying and encouraging responsible algorithmic development and deployment.

\smallsec{Identifying and addressing bias in datasets} There are several issues raised in the conversation around dataset bias. The first common issue is the lack of transparency around dataset design and collection procedures. Datasheets for Datasets~\cite{gebru2018datasheets} (and, relatedly, for models~\cite{mitchell2018model}) have been proposed as solutions, encouraging dataset creators to release detailed and standardized information on the collection protocol which can be used by downstream users to assess the suitability of the dataset. Throughout this work we dive deep into understanding the data collection pipelines of ImageNet and related datasets, and consider their implications.

The second issue is the presence of ethically questionable concepts or annotations within datasets. Examples range from quantifying beauty~\cite{beautycontest} to predicting sexual orientation~\cite{wang2018homosexuality} to (arguably) annotating gender~\cite{klare2015pushing,eidinger2014age}. In Sec.~\ref{sec:synsets} (and to a lesser extent in Sec.~\ref{sec:imageability}), we consider the underlying cause for such concepts to appear in large-scale datasets and propose the first steps of a solution.

A related consideration is the ethics and privacy of the subjects depicted in these computer vision datasets. Here we refer the reader to, e.g., Whittaker et al.~\cite{AInow2018} for a recent detailed discussion as this is outside the scope of our work.

The final and perhaps best-known source of dataset bias is the imbalance of representation, e.g., the underrepresentation of demographic groups within the dataset as a whole or within individual classes. In the context of computer vision, this issue has been brought up at least in face recognition~\cite{buolamwini2018gender}, activity recognition~\cite{zhao2017men}, facial emotion recognition~\cite{rhue2018racialface}, face attribute detection~\cite{ryu2017inclusivefacenet} and image captioning~\cite{burns2018women} -- as well as more generally in pointing out the imaging bias of datasets~\cite{torralba2011unbiased}. This is not surprising as many of the images used in computer vision datasets come from Internet image search engines, which have been shown to exhibit similar biased behavior~\cite{kay2015unequal,noble2018algorithms,celis2019implicit}. In some rare cases, a dataset has been collected explicitly to avoid such influences, e.g., the Pilot Parliaments Benchmark (PPB) by Buolamwini and Gebru~\cite{buolamwini2018gender}. In Sec.~\ref{sec:attributes} we propose a strategy for balancing ImageNet across several protected attributes while considering the implications of such design (namely the concerns with annotating pictures of individual people according to these protected attributes). 

\smallsec{Responsible algorithmic development} Beyond efforts around better dataset construction, there is a large body of work focusing on the development of fair and responsible algorithms that aim to counteract the issues which may be present in the datasets. Researchers have proposed multiple fairness metrics including statistical parity~\cite{calders2009building,kamiran2009classifying,calders2010three,edwards2015censoring}, disparate impact~\cite{feldman2015certifying,zafar2015fairness}, equalized odds~\cite{hardt2016equality} and individual fairness ~\cite{dwork2012fairness}, and 
analyzed the relationship between them~\cite{kleinberg2016inherent,pleiss2017fairness}. Algorithmic solutions have ranged from removing undesired bias by preprocessing the data~\cite{pedreshi2008discrimination,kamiran2009classifying}, striking a tradeoff between performance and fairness by posing additional regularization during training or inference~\cite{kamishima2011fairness,zemel2013learning,zafar2015fairness,hardt2016equality, zhao2017men,madras2018learning}, or designing application-specific interventions (such as of Burns et al.~\cite{burns2018women} for reducing gender bias in image captioning models).

However, statistical machine learning models have three fundamental limitations that need to be considered. First, the accuracy of a machine learning model is strongly influenced by the number of training examples: underrepresented categories in datasets will be inherently more challenging for the model to learn~\cite{liu2019large}. Second, machine learning models are statistical systems that aim to make accurate predictions on the majority of examples; this focus on common-case reasoning encourages the models to ignore some of the diversity of the data and make simplifying assumptions that may \emph{amplify} the bias present in the data~\cite{zhao2017men}. Finally, learning with constraints is a difficult open problem, frequently resulting in satisfying fairness constraints at the expense of overall model accuracy~\cite{kamiran2009classifying,kamishima2011fairness,zemel2013learning,zafar2015fairness}. Thus, algorithmic interventions alone are unlikely to be the most effective path toward fair machine learning, and dataset interventions are necessary. Even more so, most algorithmic approaches are supervised and require the protected attributes to be explicitly annotated, again bringing us back to the need for intervention at the dataset level.

\smallsec{Datasets, algorithms and intention} Finally, we note that prior work in this space underscores a single important point: any technical fairness intervention will only be effective when done in the context of the broader awareness, intentionality and thoughtfulness in building applications. Poorly constructed datasets may introduce unnoticed bias into models. Poorly designed algorithms may exploit even well-constructed datasets. Accurate datasets and models may be used with malicious intent. The responsibility for downstream fair systems lies at all steps of the development pipeline.

\section{Background: the ImageNet data collection pipeline}
\label{sec:pipeline}

To lay the groundwork for the rest of the paper, we begin by summarizing the data collection and annotation pipeline used in ImageNet as originally described in~\cite{deng2009imagenet,russakovsky2015imagenet}. This section can be safely skipped for readers closely familiar with ImageNet and related computer vision datasets, but we provide it here for completeness. 

The goal of ImageNet is to illustrate English nouns with a large number of high-resolution carefully curated images as to ``foster more sophisticated and robust models and algorithms to index, retrieve, organize and interact with images and multimedia data''~\cite{deng2009imagenet}. We consider the entire ImageNet dataset consisting of 14,197,122 images illustrating 21,841 concepts rather than just the 1,431,167 images illustrating 1,000 concepts within the ImageNet challenge which are most commonly used~\cite{russakovsky2015imagenet}. There are three steps to the ImageNet data collection pipeline: (1) selecting the concept vocabulary to illustrate, (2) selecting the candidate images to consider for each concept, and (3) cleaning up the candidates to ensure that the images in fact correspond to the target concept. We describe each step and its similarities to the steps in other vision datasets.

\smallsec{(1) Concept vocabulary} When building a visual recognition benchmark, the first decision is settling on a concept vocabulary and decide which real-world concepts should be included. WordNet~\cite{miller1998wordnet} emerges as a natural answer. It is a language ontology in which English nouns are grouped into sets of synonyms (synsets) that represent distinct semantic concepts.\footnote{We use the words ``concept'' and ``synset'' interchangeably throughout the paper.} The synsets are then organized into a hierarchy according to the ``is a'' relation, such as ``\texttt{coffee table} is a \texttt{table}''. WordNet serves as the concept vocabulary for ImageNet, which provides images for grounding the synsets visually. Similarly, subsets of the WordNet backbone have been used in datasets like Places~\cite{zhou2017places}, Visual Genome~\cite{krishna2017visual} and ShapeNet~\cite{chang2015shapenet}.

\smallsec{(2) Candidate images} The natural and easiest-to-access source of visual data is the Internet. For every concept in WordNet, the ImageNet creators queried image search engines and aimed to increase the variety of retrieved images through using multiple search engines, employing query expansion, and translating the search terms into multiple languages~\cite{deng2009imagenet}. Similarly, the vast majority of  vision datasets rely on images collected from the Internet~\cite{torralba2008tiny,krishna2017visual,zhou2017places,kuznetsova2018open,lin2014microsoft,everingham2010pascal}, with many collected by first defining the set of target concepts and then obtaining the associated images using query expansion: e.g., PASCAL~\cite{everingham2010pascal}, COCO~\cite{lin2014microsoft}, Places~\cite{zhou2017places}. 

\smallsec{(3) Manual cleanup} As noted in Torralba et al.~\cite{torralba2008tiny}, image search engines were only about 10\% accurate, and thus a manual cleanup of the candidate images is needed. The cleanup phase of ImageNet consists of a set of manual annotation tasks deployed on the Amazon Mechanical Turk (AMT) marketplace. The workers are provided with a single target concept (e.g., \texttt{Burmese cat}), its definition from WordNet, a link to Wikipedia, and a collection of candidate images. They are instructed to click on all images that contain the target concept, irrespective of any occlusion, scene clutter, or the presence of other concepts. Images that reach a desired level of positive consensus among workers are added to ImageNet. Similarly, most computer vision datasets rely on manual annotation although the details change: e.g., PASCAL was annotated in-house rather than using crowdsourcing~\cite{everingham2010pascal}, Places relies on both positive and negative verification~\cite{zhou2017places}, COCO favors very detailed annotation per image~\cite{lin2014microsoft}, and Open Images~\cite{kuznetsova2018open} and Places~\cite{zhou2017places} both use a computer-assisted annotation approach.

\smallsec{Outline} In the following sections, we consider the fairness issues that arise as a result of this pipeline, and propose the first steps to mitigate these concerns.

\section{Problem 1: Stagnant Concept Vocabulary}
\label{sec:synsets}

The backbone of WordNet~\cite{miller1998wordnet} provides a list of synsets for ImageNet to annotate with images. However, born in the past century, some synsets in WordNet are no longer appropriate in the modern context. People have reported abusive synsets in the WordNet hierarchy, including racial and sexual slurs (e.g., synsets like n10585077 and n10138472).\footnote{Throughout the paper, we refrain from explicitly listing the offensive concepts associated with synsets and instead report only their synset IDs. For a conversion, please see \url{wordnet.princeton.edu/documentation/wndb5wn}. } This is especially problematic within the \texttt{person} subtree of the concept hierarchy (i.e., synset n00007846 and its descendants). During the construction of ImageNet in 2009, the research team removed any synset explicitly denoted as ``offensive'', ``derogatory,'' ``pejorative,'' or ``slur'' in its gloss, yet this filtering was imperfect and still resulted in inclusion of a number of synsets that are offensive or contain offensive synonyms. Going further, some synsets may not be inherently offensive but may be inappropriate for inclusion in a visual recognition dataset. This filtering of the concept vocabulary is the first problem that needs to be addressed.

\subsection{Prior work on annotating offensiveness}
Sociolinguistic research has explored the problem of offensiveness, largely focusing on studying profanity. Ofcom~\cite{sueyoshi2005role} and Sapolsky et al.~\cite{sapolsky2010rating} rate the offensiveness of words in TV programs. Dewaele~\cite{dewaele2016thirty} demonstrates offensiveness to be dependent on language proficiency by studying the ratings from English native speakers and non-native speakers. Beers F{\"a}gersten~\cite{beers2007sociolinguistic} designs two questionnaires to study the role of context in the level of offensiveness of profanity. In one questionnaire, subjects see a word together with a short dialogue in which the word appears. They are asked to rate the word on a 1-10 scale from ``not offensive'' to ``very offensive''. In another questionnaire, the subjects see only the words without any context. The findings highlight the importance of context, as the perceived offensiveness depends heavily on the dialogue and on the gender and race of the subjects.

\subsection{Methodology for filtering the unsafe synsets}

We ask annotators to flag a synset as unsafe when it is either inherently ``offensive,'' e.g., containing profanity or racial or gender slurs, or ``sensitive,'' i.e., not inherently offensive but may cause offense when applied inappropriately, such as the classification of people based on sexual orientation and religion. In contrast to prior work, we do not attempt to quantify the ``level'' of offensiveness of a concept but rather exclude all potentially inappropriate synsets. Thus, we do not adopt a 1-5 or 1-10 scale~\cite{dewaele2016thirty, beers2007sociolinguistic} for offensiveness ratings. Instead, we instruct workers to flag any synset that may be potentially unsafe, essentially condensing the rating 2-5 or 2-10 on the scale down to a single ``unsafe'' label.

\subsection{Results and impact on ImageNet after removing the unsafe synsets}
We conduct the initial annotation using in-house workers, who are 12 graduate students in the department and represent 4 countries of origin, male and female genders, and a handful of racial groups. The instructions are available in \hyperref[appendix:1]{Appendix}. So far out of 2,832 synsets within the \texttt{person} subtree, we have identified 1,593 unsafe synsets. The remaining 1,239 synsets are temporarily deemed safe. Table~\ref{table:offensiveness} gives some examples of the annotation results (with the actual content of offensive synsets obscured). The full list of synset IDs can be found in \hyperref[appendix:1]{Appendix}. The unsafe synsets are associated with 600,040 images in ImageNet. Removing them would leave 577,244 images in the safe synsets of the \texttt{person} subtree of ImageNet.  

\subsection{Limitations of the offensiveness annotation and future work }

First, it is important to note that a ``safe'' synset only means that the label itself is not deemed offensive. It does not mean that it is possible, useful, or ethical to infer such a label from visual cues.

Second, despite the preliminary results we have, offensiveness is subjective and also constantly evolving, as terms develop new cultural context. Thus, we are opening up this question to the community. We are in the process of updating the ImageNet website to allow users to report additional synsets as unsafe. While the dataset may be large scale, the number of remaining concepts is  relatively small (here in the low thousands and further reduced in the next section), making this approach feasible.

\begin{table*}[t]
\caption{Examples of synsets in the \texttt{person} subtree annotated as unsafe (offensive), unsafe (sensitive), safe but non-imageable, and simultaneously safe and imageable. For unsafe (offensive) synsets, we only show their synset IDs. The annotation procedure for distinguishing between unsafe and safe synsets is described in Sec.~\ref{sec:synsets}; the procedure for non-imageable vs. imageable is in Sec.~\ref{sec:imageability}. We recommend removing the synsets of the first two columns from ImageNet entirely, and refrain from using synsets from the third column when training visual recognition models.}
\label{table:offensiveness}
\begin{center}
\begin{small}
\begin{tabular}{llll}
\toprule
Unsafe (offensive) & Unsafe (sensitive) & Safe non-imageable & {\bf Safe imageable} \\
\midrule
\texttt{n10095420: <sexual slur>} & \texttt{n09702134: Anglo-Saxon} & \texttt{n10002257: demographer} & \texttt{n10499631: Queen of England} \\ 
\texttt{n10114550: <profanity>} & \texttt{n10693334: taxi dancer} & 
\texttt{n10061882: epidemiologist} & \texttt{n09842047: basketball player} \\  
\texttt{n10262343: <sexual slur>} & \texttt{n10384392: orphan} &\texttt{n10431122: piano maker} & \texttt{n10147935: bridegroom} \\ 
\texttt{n10758337: <gendered slur>} & \texttt{n09890192: camp follower} & \texttt{n10098862: folk dancer} & \texttt{n09846755: beekeeper} \\
\texttt{n10507380: <criminative>} & \texttt{n10580030: separatist} & \texttt{n10335931: mover} & \texttt{n10153594: gymnast} \\  
\texttt{n10744078: <criminative>} &  \texttt{n09980805: crossover voter} & \texttt{n10449664: policyholder} & \texttt{n10539015: ropewalker} \\ 
\texttt{n10113869: <obscene>} & \texttt{n09848110: theist} & \texttt{n10146104: great-niece} & \texttt{n10530150: rider} \\ 
\texttt{n10344121: <pejorative>} & \texttt{n09683924: Zen Buddhist} & \texttt{n10747119: vegetarian} & \texttt{n10732010: trumpeter} \\ 
\bottomrule
\end{tabular}
\end{small}
\end{center}
\end{table*}

\section{Problem 2: Non-visual Concepts}
\label{sec:imageability}

ImageNet attempts to depict each WordNet synset with a set of images. However, not all synsets can be characterized visually. For example, is it possible to tell whether a person is a \texttt{philanthropist} from images? This issue has been partially addressed in ImageNet's annotation pipeline~\cite{deng2009imagenet} (Sec.~\ref{sec:pipeline}), where candidate images returned by search engines were verified by human annotators.  It was observed that different synsets need different levels of consensus among annotators, and a simple adaptive algorithm was devised to determine the number of agreements required for images from each synset. Synsets harder to characterize visually would require more agreements, which led to fewer (or no) verified images.

Despite the adaptive algorithm, we find a considerable number of the synsets in the \texttt{person} subtree of ImageNet to be non-imageable---hard to characterize accurately using images. There are several reasons. One reason for them sneaking into ImageNet could be the large-scale annotation. Although non-imageable synsets require stronger consensus and have fewer verified images, they remain in ImageNet as long as there are some images successfully verified, which is likely given the large number of images. Another reason could be ``positive bias'': annotators are inclined to answer ``yes'' when asked the question ``Is there a \texttt{<concept>} in the image?'' As a result, some images with weak visual evidence of the corresponding synset may be successfully verified.

The final and perhaps most compelling reason for non-imageable synsets to have been annotated in ImageNet is that search engines will surface the most \emph{distinctive} images for a concept, even if the concept itself is not imageable. For example, identifying whether someone is \texttt{Bahamian} from a photograph is not always be possible, but there will be some distinctive images (e.g., pictures of a people wearing traditional Bahamian costumes), and those will be the ones returned by the search engine.
This issue is amplified by the presence of stock photography on the web, which contributes to and perpetuates stereotypes as discussed at length in e.g., ~\cite{frosh2001inside,frosh2002rhetorics,aiello2016corporations}. Overall, this results in an undoubtedly biased visual representation of the categories, and while the issue affects all synsets, it becomes particularly blatant for categories that are inherently non-imageable. Thus in an effort to reduce the visual bias, we explicitly determine the imageability of the synsets in the \texttt{person} subtree and recommend that the community refrain from using those with low imageability when training visual recognition models.

\subsection{Annotating imageability} 
Extensive research in psycholinguistics has studied the imageability (a.k.a. imagery) of words~\cite{paivio1965abstractness,paivio1968concreteness,gilhooly1980age,altarriba1999concreteness,bird2001age,cortese2004imageability}, which is defined as ``the ease with which the word arouses imagery''~\cite{paivio1968concreteness}.
For annotating imageability, most prior works follow a simple procedure proposed by Paivio et al.~\cite{paivio1968concreteness}:
The workers see a list of words and rate each word on a 1-7 scale from ``low imagery (1)''  to ``high imagery (7)''. For each word, the answers are averaged to establish the final score.

We adopt this definition of imageability and adapt the existing procedure to annotate the imageability of synsets in the ImageNet \texttt{person} subtree. However, unlike prior works that use in-house workers to annotate imageability, we rely on crowdsourcing. This allows us to scale the annotation and obtain ratings from a diverse pool of workers~\cite{ross2009turkers,difallah2018demographics}, but also poses challenges in simplifying the instructions and in implementing robust quality control. 

In our crowdsourcing interface (in \hyperref[appendix:2]{Appendix}), we present the human subject with a list of concepts and ask them to identify how easy it is to form a mental image of each. To reduce the cognitive load on the workers, we provide a few examples to better explain the task, include the synonyms and the definition of each concept from WordNet, and change the rating to be a simpler 5-point (rather than 7-point) scale from ``very hard (1)'' to ``very easy (5)''. The final imageability score of a synset is an average of the ratings. 

For quality control, we manually select 20 synsets as gold standard questions (in \hyperref[appendix:2]{Appendix}); half of them are obviously imageable (should be rated 5), and the other half are obviously non-imageable (should be rated 1). They are used to evaluate the quality of workers. If a worker has a high error on the gold standard questions, we remove all the ratings from this worker. We also devise a heuristic algorithm to determine the number of ratings to collect for each synset. Please refer to \hyperref[appendix:2]{Appendix} for details.

\begin{figure}[t]
\centering
\includegraphics[width=1.0\columnwidth]{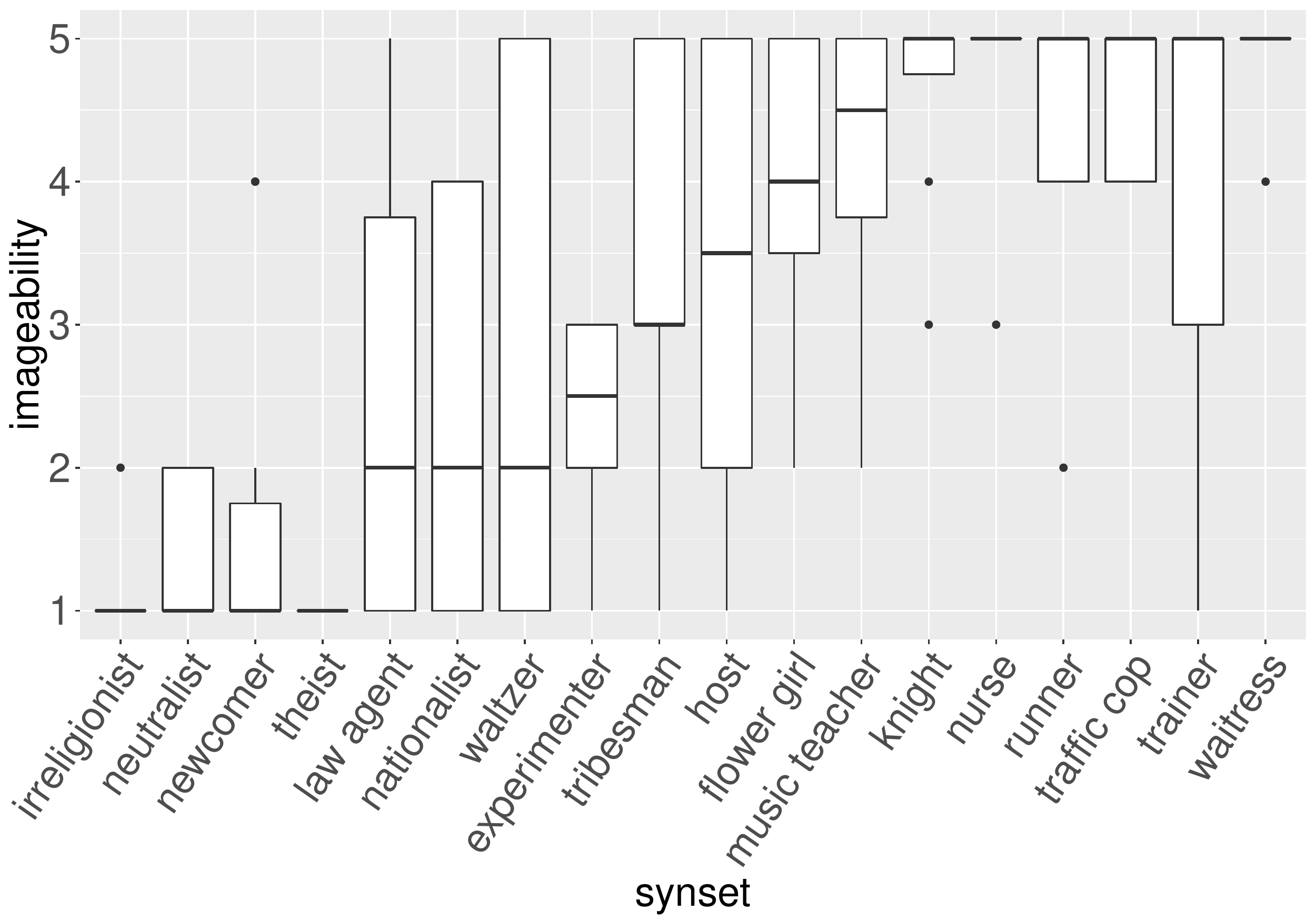}
\caption{The distribution of raw imageability ratings for selected synsets. \texttt{irreligionist} and \texttt{nurse} have more well-accepted imageability than \texttt{host} and \texttt{waltzer}.}
\label{fig:imageability_synsets}
\end{figure}

\begin{figure}[t]
\centering
\includegraphics[width=1.0\columnwidth]{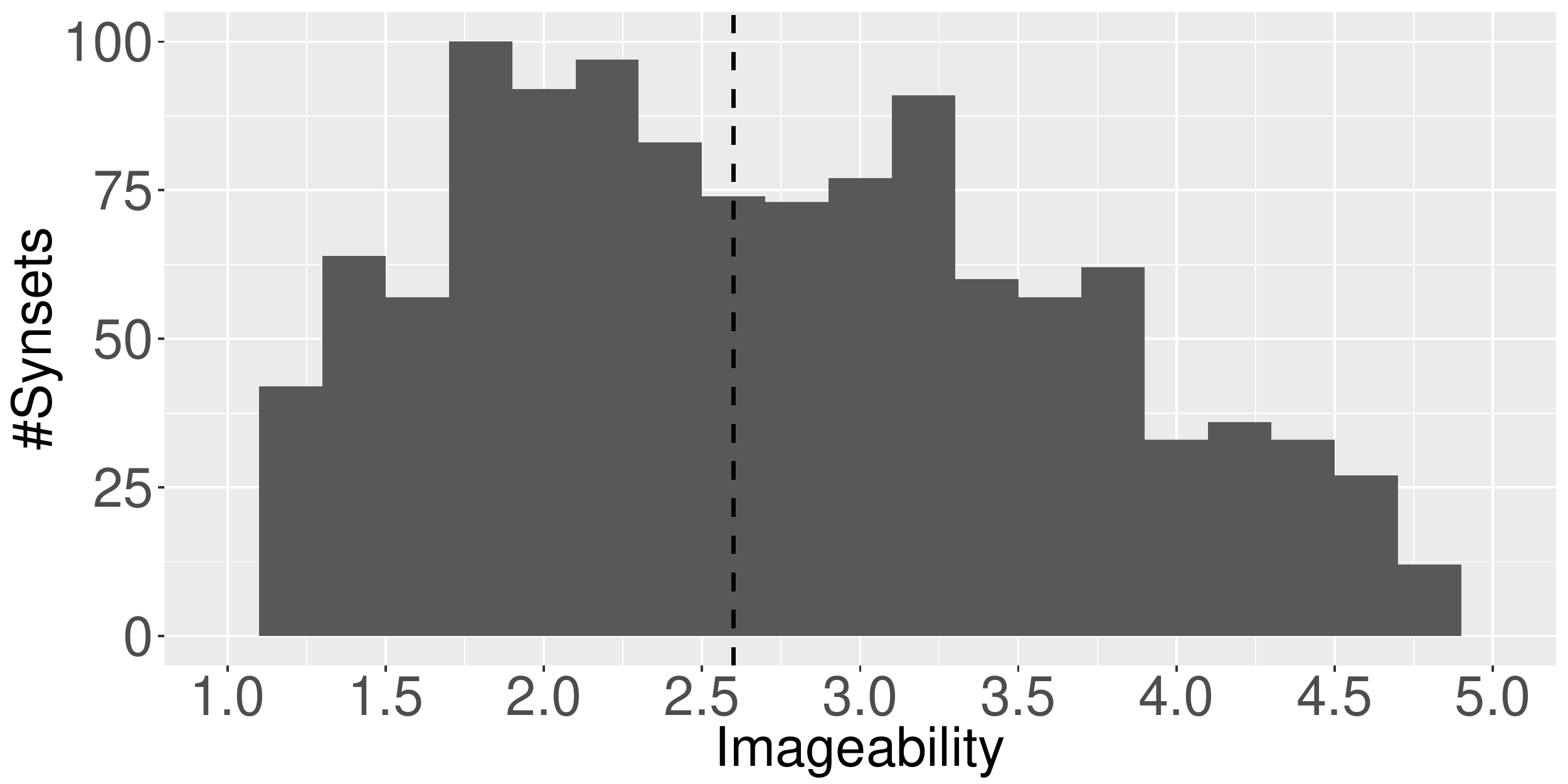}
\caption{The distribution of the final imageability scores for all of the 1,239 safe synsets. The median is 2.60.}
\label{fig:imageability_hist}
\end{figure}

\subsection{Results and impact on ImageNet after removing the non-imageable synsets}
We annotate the imageability of 1,239 synsets in the \texttt{person} subtree which have been marked as safe synsets in the previous task. Fig.~\ref{fig:imageability_synsets} shows the imageability ratings for a selected set of synsets. Synsets such as \texttt{irreligionist} and \texttt{nurse} have well-accepted imageability (\texttt{irreligionist} is deemed to be decidedly non-imageable, \texttt{nurse} is deemed to be clearly imageable). In contrast, it is much harder to reach a consensus on the imageability of \texttt{host} and \texttt{waltzer}. Fig.~\ref{fig:imageability_hist} shows the distribution of the final imageability scores for all of the 1,239 safe synsets. The median is 2.60; only 158 synsets have imageability greater than or equal to 4. Table~\ref{table:offensiveness} shows some examples of non-imageable synsets. The complete list is in \hyperref[appendix:2]{Appendix}.

After manually examining the results, we suggest that all synsets in the \texttt{person} subtree with imageability less than 4 be considered ``non-imageable'' and not be used for training models. There would be 443,547 images and 1,081 synsets flagged, including \texttt{hobbyist} (1.20), \texttt{job candidate} (2.64), and \texttt{bookworm} (3.77); there would be 133,697 images and 158 synsets remaining, including \texttt{rock star} (4.86), \texttt{skier} (4.50), and \texttt{cashier} (4.20). More examples are in Table~\ref{table:offensiveness}. Future researchers are free to adjust this threshold as needed. 
 
\subsection{Limitations of the imageability annotation} By manually examining a subset of the synsets, we find the imageability results to be reasonable overall, but we also observe a few interesting exceptions. Some synsets with high imageability are actually hard to characterize visually, e.g., \texttt{daughter} (5.0) and \texttt{sister} (4.6); they should not have any additional visual cues besides being female. Their high imageability scores could be a result of the mismatch between ``the ease to arouse imagery'' and ``the ease to characterize using images''. \texttt{Daughter} and \texttt{sister} are hard to characterize visually, but they easily arouse imagery if the annotator has a daughter or a sister. The definition based on ease of characterization with visual cues is more relevant to computer vision datasets, but we adopt the former definition as a surrogate since it is well-accepted in the literature, and there are mature procedures for annotating it using human subjects.

Another interesting observation is that workers tend to assign low imageability to unfamiliar words. For example, \texttt{cotter} (a peasant in the Scottish Highlands) is scored 1.70 while the generic \texttt{peasant} is scored 3.36. Prior works have demonstrated a strong correlation between familiarity and imageability~\cite{gilhooly1980age, bird2001age, yee2017valence}, which explains the low imageability of the less frequent \texttt{cotter}. However, low familiarity with a concept is anyway an important factor to consider in crowdsourcing dataset annotation, as unfamiliar terms are more likely to be misclassified by workers. This suggests that removing synsets identified as less imageable by our metric may also have the additional benefit of yielding a more accurate dataset.\footnote{Further, unfamiliar terms are also likely to be less common and thus less relevant to the downstream computer vision task, making their inclusion in the dataset arguably less important.}

When analyzing Table~\ref{table:offensiveness}, we further wonder whether even the synsets that are both safe and imageable should remain in ImageNet. For example, is the \texttt{Queen of England} an acceptable category for visual recognition? Would \texttt{basketball player} be better replaced with \texttt{person interacting with a basketball} and captured as a human-object-interaction annotation? Would \texttt{bridegroom} be rife with cultural assumptions and biases? As always, we urge downstream users to exercise caution when training on the dataset.

And finally, we observe that even the remaining imageable synsets may contain biased depictions as a result of search engine artifacts. For example, the synset \texttt{mother} (imageability score 4.3) primarily contains women holding children; similarly, the synset \texttt{beekeeper} (imageability score 4.6) predominantly contains pictures of people with bees. Even though one remains a mother when not around children, or a beekeeper when not around bees, those images would rarely be surfaced by a search engine (and, to be fair, would be difficult for workers to classify even if they had been).

Despite these concerns about the imageability annotations and about the lingering search engine bias, one thing that is clear is that at least the non-imageable synsets are problematic. Our annotation of imageability is not perfect and not the final solution, but an important step toward a more reasonable distribution of synsets in the ImageNet \texttt{person} subtree.

\subsection{Relationship between imageability and visual recognition models} \label{sec:accuracy} To conclude the discussion of imageability, we ask one final question: What is the relationship between the imageability of synset and the accuracy of a corresponding visual recognition model? Concretely, are the imageable synsets actually easier to recognize because they correspond to visual concepts?  Or, on the flip side, is it perhaps always the case that non-imageable synsets contain an overly-simplified stereotyped representation of the concept and thus are easy for models to classify? If so, this would present additional evidence about the dangers of including such categories in a dataset since their depicted stereotypes are easily learned and perpetuated by the models.

\smallsec{Computer vision experiment setup} To evaluate this, we run a simple experiment to study the relationship between the imageability of a synset and the ability of a modern deep learning-based image classifier to recognize it. We pick a subset of 143 synsets from the 1,239 safe synsets so that each synset has at least 1,000 images. The selected synsets are leaf nodes in the WordNet hierarchy, meaning that they cannot be ancestors of each other and they represent disjoint concepts. We randomly sample 1,000 images from each synset, 700 for training, 100 for validation, and 200 for testing. We use a standard  ResNet34 network~\citep{he2016deep} to classify the images as belonging to one of the 143 synsets. During training, the images are randomly cropped and resized to $224 \times 224$; we also apply random horizontal flips. During validation and testing, we take $224 \times 224$ crops at the center. The network is trained from scratch for 90 epochs, which takes two days using a single GeForce GTX 1080 GPU. We minimize the cross-entropy loss using stochastic gradient descent; the learning rate starts at 0.05 and decreases by a factor of 10 every 30 epochs. We also use a batch size of 256, a momentum of 0.9, and a weight decay of 0.001.

\begin{figure*}[t]
\centering
\includegraphics[width=2.0\columnwidth]{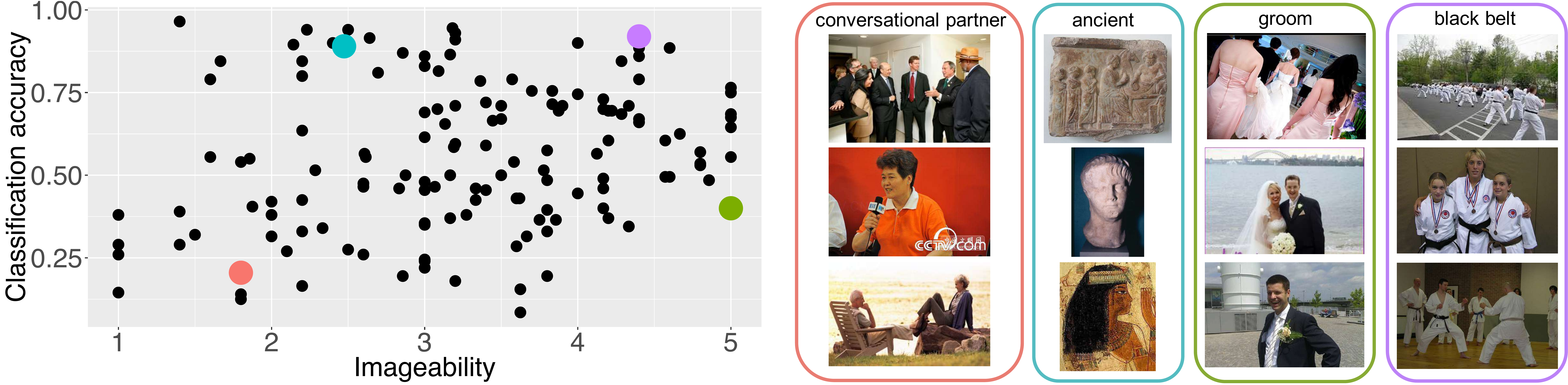}
\caption{(Left) The computer vision model's classification accuracy vs. synset imageability for 143 safe synsets which contain at least 1000 images. More imageable synsets are not necessarily easier for models to recognize, with Pearson correlation coefficient
 $r=0.23$. (Right) Example images from synsets that are non-imageable and hard to classify (\texttt{conversational partner}); non-imageable but easy to classify (\texttt{ancient}); imageable but hard to classify (\texttt{groom}); imageable and easy to classify (\texttt{black belt}).}
\label{fig:accuracy_imageability}
\end{figure*}

\smallsec{Computer vision experiment results} The network has an overall testing accuracy of 55.9\%. We are more interested in the breakdown accuracies for each synset and how they correlate with the imageability. The network's testing accuracy on the easily imageable synsets (score $\geq$ 4) is 63.8\%, which is higher than the accuracy of 53.0\% on the synsets deemed non-imageable (score $<$ 4). Overall there is a positive correlation between imageability and accuracy (Pearson correlation coefficient $r=0.23$ with a p-value of $0.0048$) as depicted in Fig.~\ref{fig:accuracy_imageability} (left). To better understand this, we analyze four representative examples, also depicted in Fig.~\ref{fig:accuracy_imageability} (right), which highlight the different aspects at play here:

\begin{itemize}
   \item {\it Imageable, easy to classify:} A category such as \texttt{black belt} is both deemed imageable (score of 4.4) and is easy to classify (accuracy of 92\%). The retrieved images contain visually similar results that are easy to learn by the model and easy to distinguish from other people categories. 
    \item {\it Non-imageable, hard to classify:} On the other end of the spectrum, \texttt{conversational partner} is deemed non-imageable (score of only 1.8) as it doesn't evoke a prototypical visual example. The images retrieved from search engines contain groups of people engaged in conversations, so the annotators verifying these images in the ImageNet pipeline correctly labeled these images as containing a \texttt{conversation partner}. However, the resulting set of images is too diverse, and the visual cues are too weak to be learned by the model (accuracy only 20.5\%) . 
    \item {\it Imageable, hard to classify:} \texttt{Bridegroom} (synset ID n10147935) is an example of a category with a mismatch between imageability and accuracy. It is annotated as imageable (perfect score of 5.0), because it easily arouses imagery (albeit highly culturally-biased imagery). The retrieved search result images are as expected culturally biased, but correctly verified for inclusion in ImageNet. However, the accuracy of the classifier in this case is low (only 40\%) partially because of the visual diversity of the composition of images but primarily because of confusion with a closely related synset n10148035, which also corresponds to the term \texttt{bridegroom} but with a slightly different definition (n10147935: a man who has recently been married, versus n10148035: a man participant in his own marriage ceremony). This highlights the fact that classification accuracy is not a perfect proxy for visual distinctiveness, as it depends not only on the intra-synset visual cues but also on the inter-synset variability.
    \item {\it Non-imageable, easy to classify:} Finally,  \texttt{Ancient} (person who lived in ancient times) is deemed non-imageable (score of 2.5), because the imageability annotators have never seen such a person, so it is difficult to properly imagine what they might look like. However, the image search results are highly biased to ancient artifacts, including images that are not even people. The annotators agreed that these images correspond to the word \texttt{ancient}, at times making mistakes in failing to read the definition of the synset and annotating ancient artifacts as well. In the resulting set of images, visual classifiers would have no difficulty distinguishing this set of images with distinctive color patterns and unusual objects from the other people categories (accuracy 89\%). 
\end{itemize}

The findings highlight the intricacies of image search engine results, of the ImageNet annotation pipeline, of the imageability annotations, and of evaluating visual distinctiveness using visual classifiers. A deeper analysis is needed to understand the level of impact of each factor, and we leave that to future research. Until then, we suggest that the community refrain from using synsets deemed non-imageable when training visual recognition models, and we will update ImageNet to highlight that.

\begin{figure*}[ht]
\centering
\includegraphics[width=0.66\columnwidth]{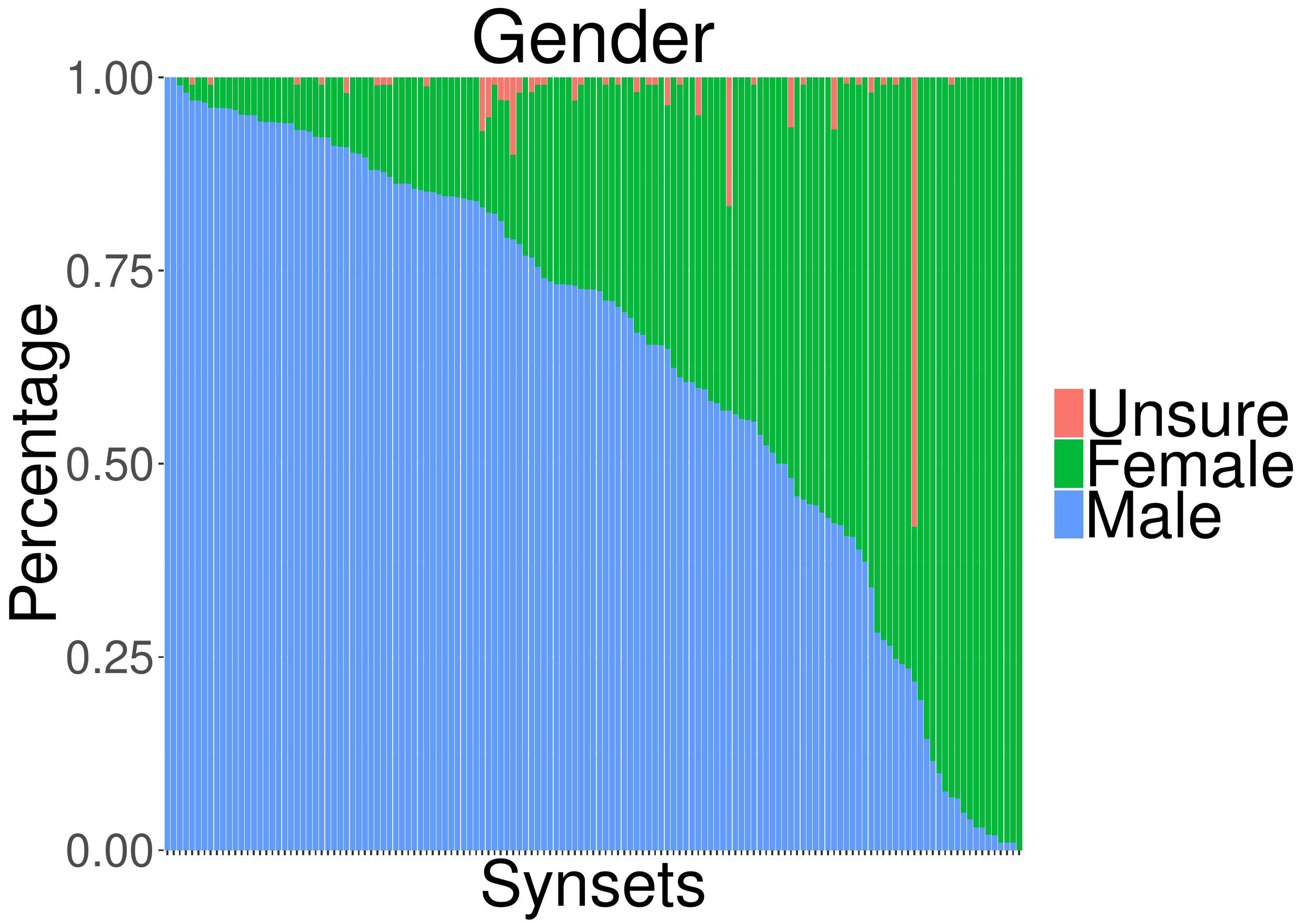}
\includegraphics[width=0.66\columnwidth]{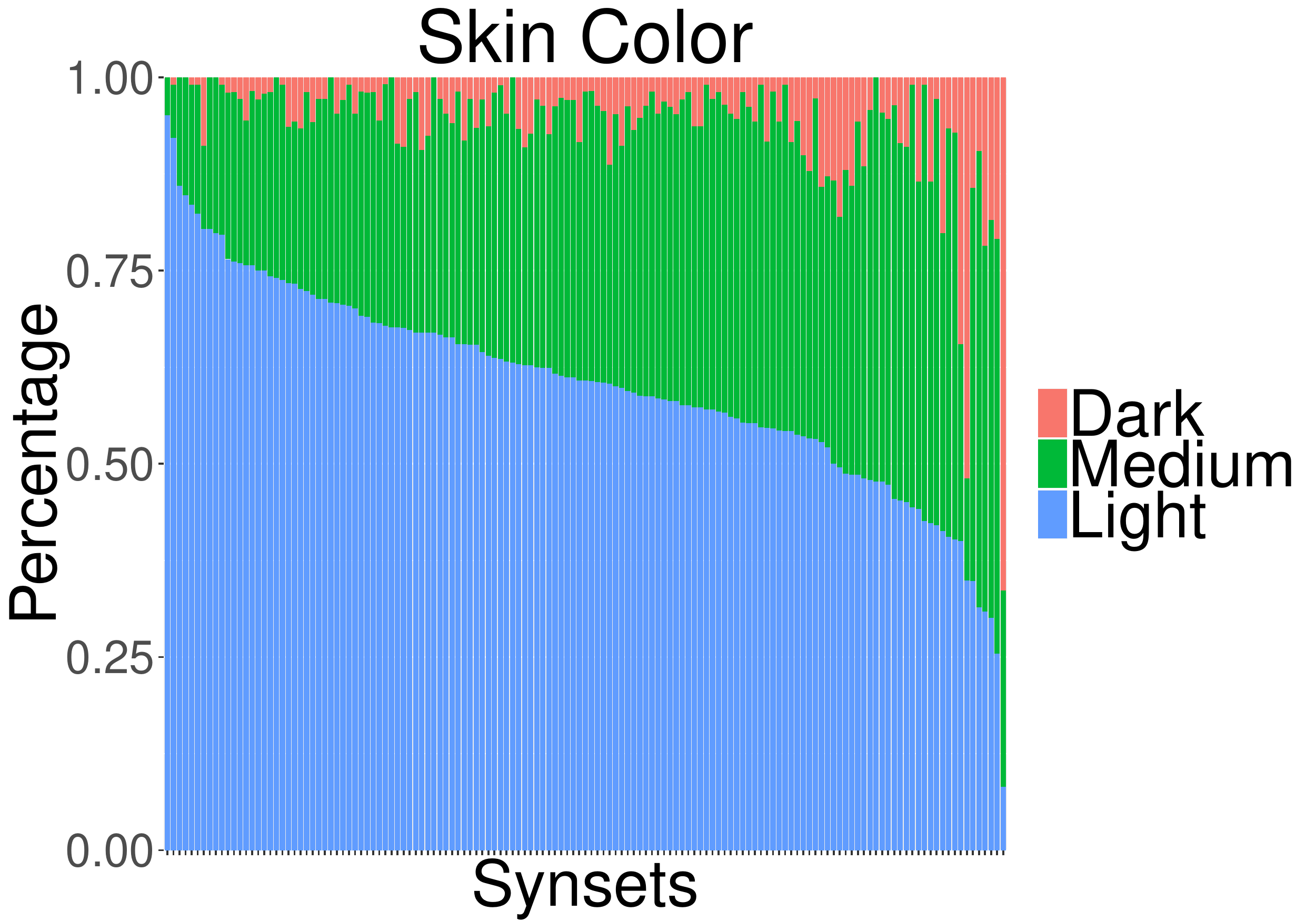}
\includegraphics[width=0.66\columnwidth]{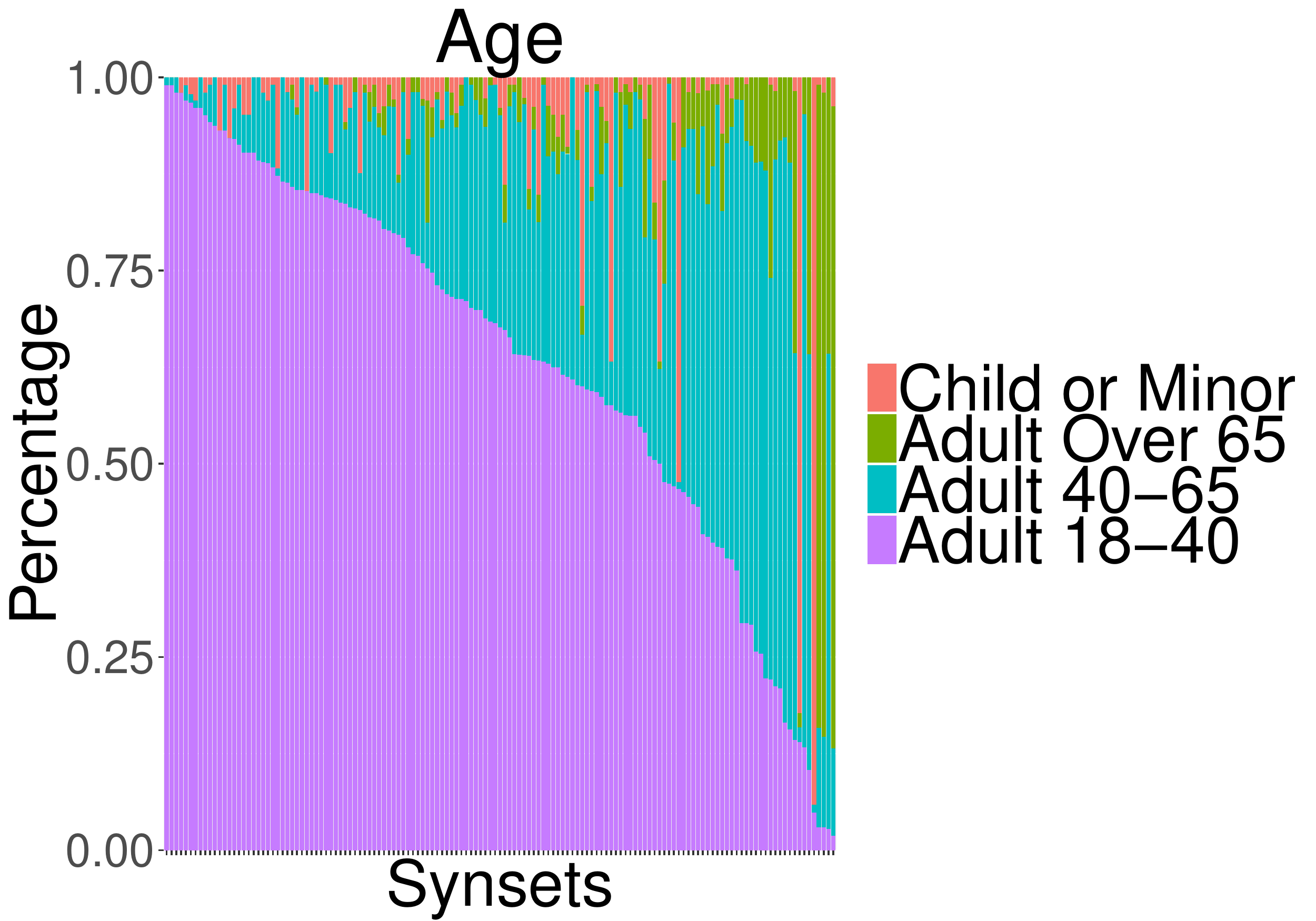}
\caption{The distribution of demographic categories across the 139 safe and imageable synsets which contain at least 100 images. The size of the different color areas reveal the underrepresentation of certain groups.}
\label{fig:attrs_results}
\end{figure*}

\section{Problem 3: Lack of Image Diversity}
\label{sec:attributes}

So far we have considered two problems: the inclusion of potentially offensive concepts (which we will remove) and the illustration of non-imageable concepts with images (which we will clearly identify in the dataset). 
The last problem we consider is insufficient representation among ImageNet images. ImageNet consists of Internet images collected by querying image search engines~\cite{deng2009imagenet}, which have been demonstrated to retrieve biased results in terms of race and gender~\cite{kay2015unequal,noble2018algorithms,celis2019implicit}. Taking gender as an example, Kay et al.~\cite{kay2015unequal} find that when using occupations (e.g., banker) as keywords, the image search results exhibit exaggerated gender ratios compared to the real-world ratios. In addition, bias can also be introduced during the manual cleanup phase when constructing ImageNet, as people are inclined to give positive responses when the given example is consistent with stereotypes~\cite{kay2015unequal}.

ImageNet has taken measures to diversify the images, such as keywords expansion, searching in multiple languages, and combining multiple search engines. Filtering out non-imageable synsets also mitigates the issue: with stronger visual evidence, the workers may be less prone to stereotypes. Despite these efforts, the bias in protected attributes remains in many synsets in the \texttt{person} subtree. It is necessary to study how this type of bias affects models trained for downstream vision tasks, which would not be possible without high-quality annotation of image-level demographics.

\subsection{Prior work on annotating demographics}
Image-level annotation of demographics is valuable for research in machine learning fairness. However, it is difficult to come up with a categorization of demographics, especially for gender and race. Buolamwini and Gebru~\cite{buolamwini2018gender} adopt a binary gender classification and the Fitzpatrick skin type classification system~\cite{fitzpatrick1988validity}. Zhao et al.~\cite{zhao2017men} and Kay et al.~\cite{kay2015unequal} also adopt a binary gender classification. Besides \textit{Male} and \textit{Female}, Burns et al.~\cite{burns2018women} add another category \textit{Neutral} to include people falling out of the binary gender classification. Ryu et al.~\cite{ryu2017inclusivefacenet} do not explicitly name the gender and race categories, but they have discrete categories nevertheless: five race categories (\textit{S1}, \textit{S2}, \textit{S3}, \textit{S4}, \textit{Other}) and three gender categories (\textit{G1}, \textit{G2}, \textit{Other}). 

\subsection{Methodology for annotating demographics} 

\smallsec{Annotated attributes} To evaluate the demographics within ImageNet and propose a more representative subset of images, we annotate a set of protected attributes on images in the \texttt{person} subtree. We consider U.S. anti-discrimination laws, which name race, color, national origin, religion, sex, gender, sexual orientation, disability, age, military history, and family status as protected attributes~\cite{epa1963law,civil1984law,ada1990law}. Of these, the only imageable attributes are color, gender, and age, so we proceed to annotate these. 

\begin{enumerate}
    \item {\it Gender.} We annotate perceived gender rather than gender identity, as someone's gender identity may differ from their gender expression and thus not be visually prominent. It is debatable what a proper categorization of gender is and whether gender can be categorized at all. Rather than addressing the full complexity of this question, we follow prior work~\cite{kay2015unequal,zhao2017men,ryu2017inclusivefacenet,burns2018women,buolamwini2018gender} and use a set of discrete categories: \textit{Male}, \textit{Female}, and \textit{Unsure}, in which \textit{Unsure} is used to both handle ambiguous visual cues as well as to include people with diverse gender expression. 
\item {\it Skin color.} We annotate skin color according to an established dermatological metric---individual typology angle (ITA)~\cite{chardon1991skin}. It divides the spectrum of skin color into 6 groups, which is too fine-grained for our purpose. Instead, we combine the groups into \textit{Light}, \textit{Medium}, and \textit{Dark}. Melanin index~\cite{takiwaki1998measurement} is another metric for skin color, which is used by the Fitzpatrick skin type classification system. However,  we opt to use the more modern ITA system. Similar to prior work~\cite{buolamwini2018gender}, skin color is used as a surrogate for race membership because it is more visually salient. 
\item {\it Age. } We annotate perceived age groups according to discrimination laws, which led to the categories of \textit{Child or Minor (under 18 years old)}, \textit{Adult (18-40)}, \textit{Over 40 Adult (40-65)}, and \textit{Over 65 Adult (65+)}. 
\end{enumerate}

\smallsec{Annotation instructions} We use crowdsourcing to annotate the attributes on Amazon Mechanical Turk. We downloaded all ImageNet images from safe synsets in the \texttt{person} subtree whose imageability score is 4.0 or higher. An image and the corresponding synset form a task for the workers, and consists of two parts. First, the worker sees the image, the synset (including all words in it), and its definition in WordNet~\cite{miller1998wordnet} and is asked to identify \emph{all} persons in the image who look like members of the synset. If at least one person is identified, the worker proceeds to annotate their gender, skin color, and age. The labels are image-level, rather than specific to individual persons. There can be multiple labels for each attribute.  For example, if the worker identified two persons in the first phase, they may check up to two labels when annotating the gender. The user interface is in \hyperref[appendix:3]{Appendix}.

The task is less well-defined when multiple persons are in the image. It can be difficult to tell which person the synset refers to, or whether the person exists in the image at all. We have tried to use automatic methods (e.g., face detectors) to detect people before manually annotating their demographics. However, the face detector is a trained computer vision model and thus also subject to dataset bias. If the face detector is only good at detecting people from a particular group, the annotation we get will not be representative of the demographic distribution in ImageNet. Therefore, we opt to let workers specify the persons they annotate explicitly. 

\smallsec{Quality control} For quality control, we have pre-annotated a set of gold-standard questions (in \hyperref[appendix:3]{Appendix}) for measuring the quality of workers. The worker's accuracy on a gold standard question $i$ is measured by intersection-over-union (IOU):
\begin{equation}
IOU_i = \frac{| A_i \cap G_i |}{| A_i \cup G_i |}
\end{equation}
where $A_i$ is the set of categories annotated by the worker, and $G_i$ is the set of ground truth categories.
For example, for an image containing a black female adult and a white female child, $G_i = \{Dark, Light, Female, Adult, Child\}$.
If a worker mistakenly take the child to be an adult and annotates $A_i = \{Dark, Light, Female, Adult\}$, the annotation quality is computed as $IOU_i = 4 / 5 = 0.8$. We exclude all responses from workers whose average $IOU$ is less than 0.5.
After removing high-error workers, we aggregate the annotated categories of the same image from independent workers. 
Each image is annotated by at least two workers. For any specific category (e.g. \textit{Adult}), we require consensus from $\max \{2, \lceil n_i / 2 \rceil \}$ workers, where $n_i$ is the number of workers for this image. For any image, we keep collecting annotations from independent workers until the consensus is reached. In the annotation results, the consensus is reached with only two workers for 70.8\% of the images; and 4 workers are enough for 97.3\% images.

\begin{figure*}
    \centering
    \begin{small}
    \begin{tabular}{c@{}c@{}c@{}c@{}c@{}c@{}c}
\multicolumn{7}{c}{Original:}\\
\includegraphics[height=0.59in]{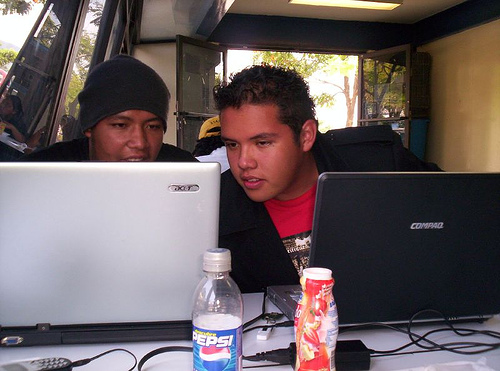} &
\includegraphics[height=0.59in]{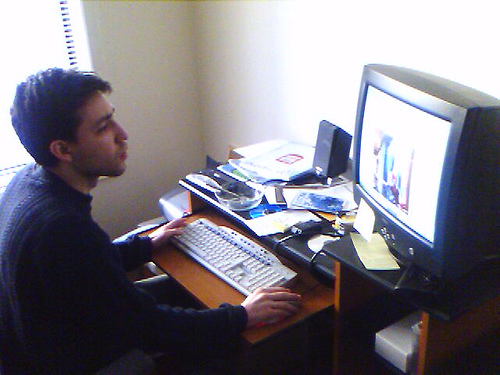} &
\includegraphics[height=0.59in]{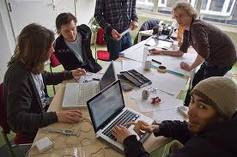} &
\includegraphics[height=0.59in]{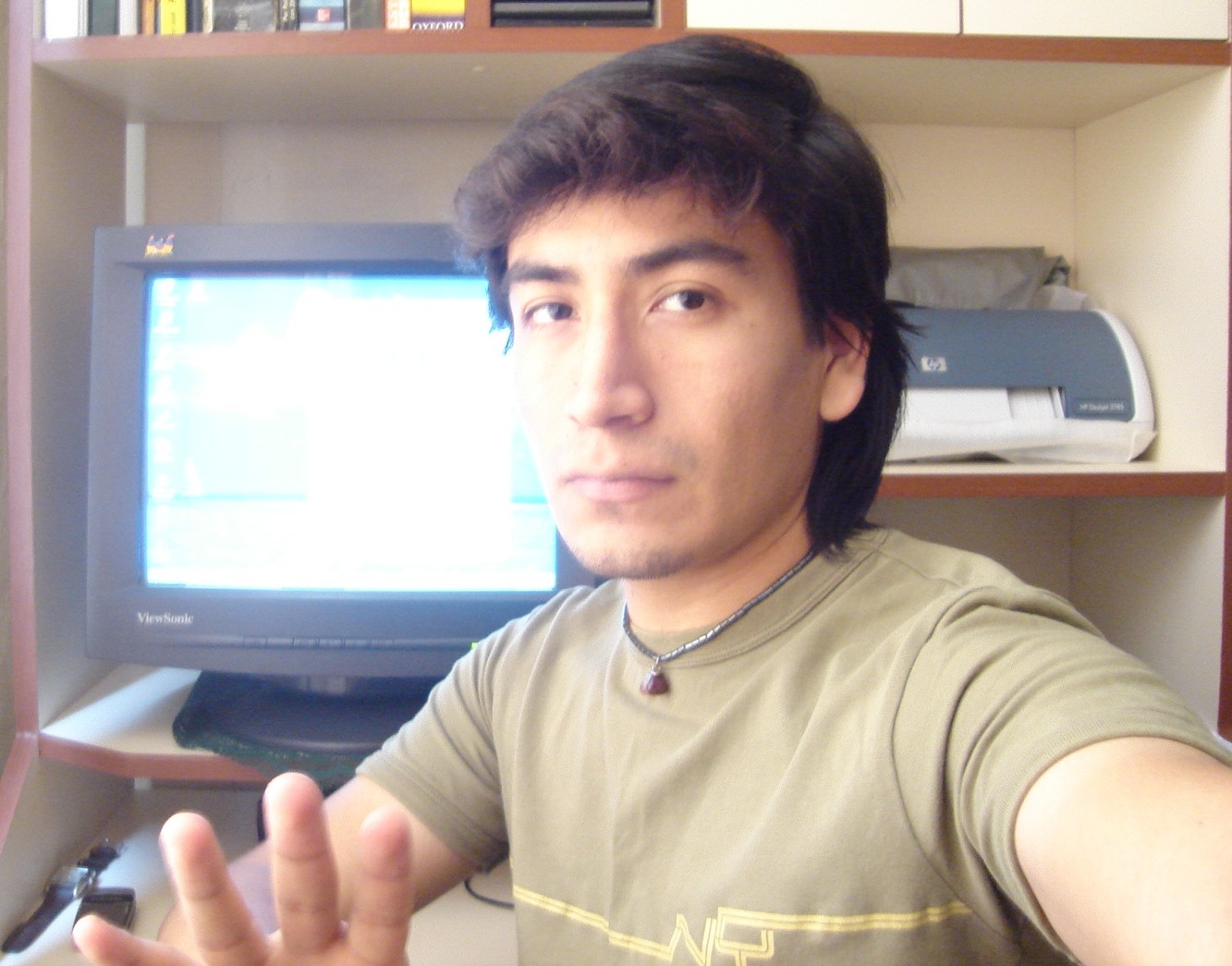} &
\includegraphics[height=0.59in]{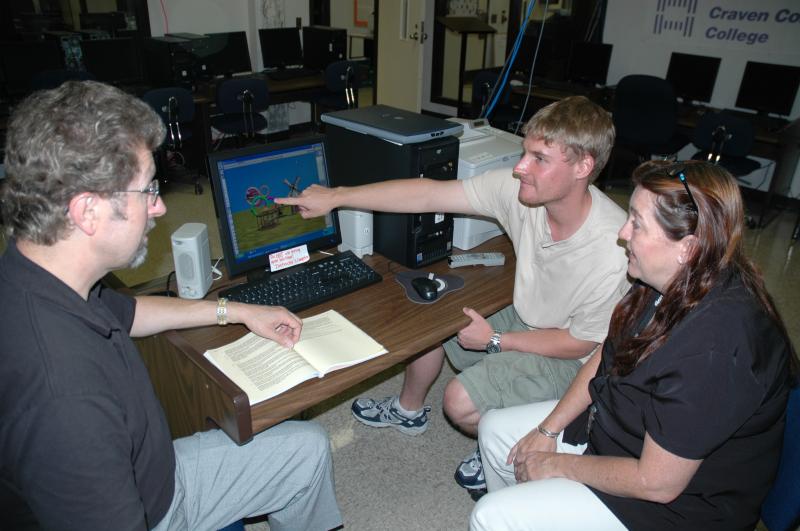} &
\includegraphics[height=0.59in]{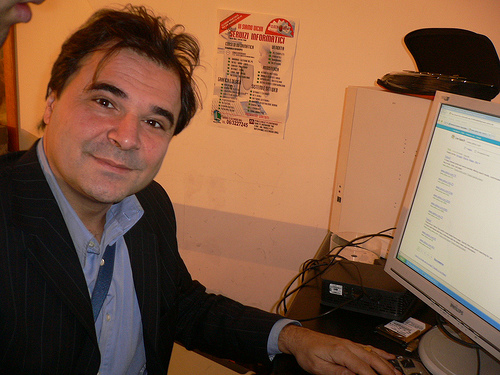} &
\includegraphics[height=0.59in]{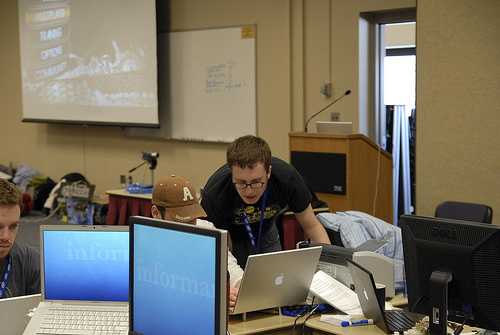} \\
\multicolumn{7}{c}{Balancing gender:}\\
\includegraphics[height=0.59in]{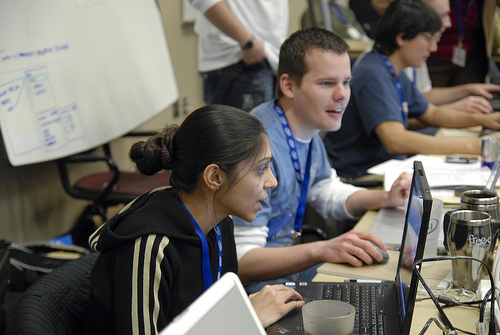} &
\includegraphics[height=0.59in]{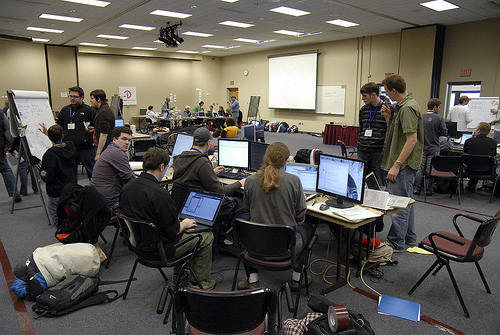} &
\includegraphics[height=0.59in]{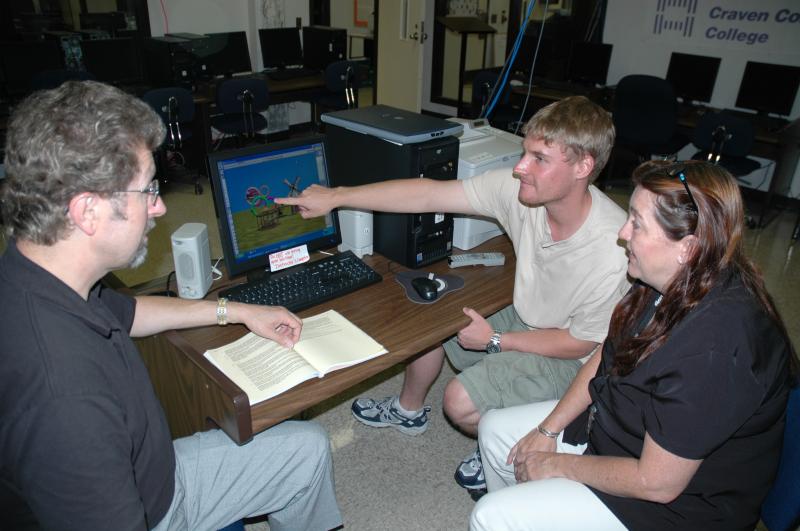} &
\includegraphics[height=0.59in]{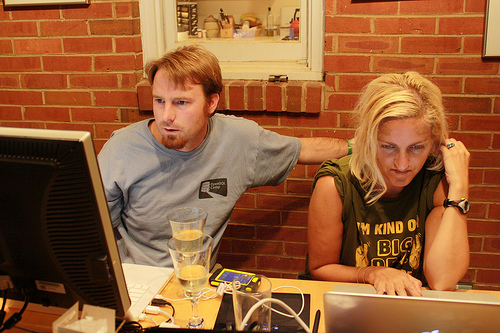} &
\includegraphics[height=0.59in]{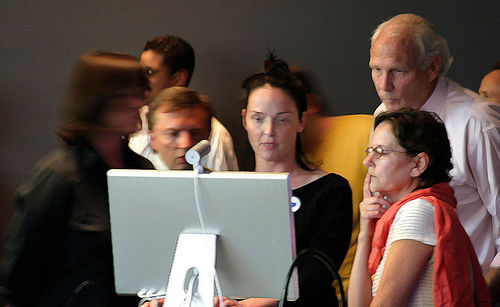} &
\includegraphics[height=0.59in]{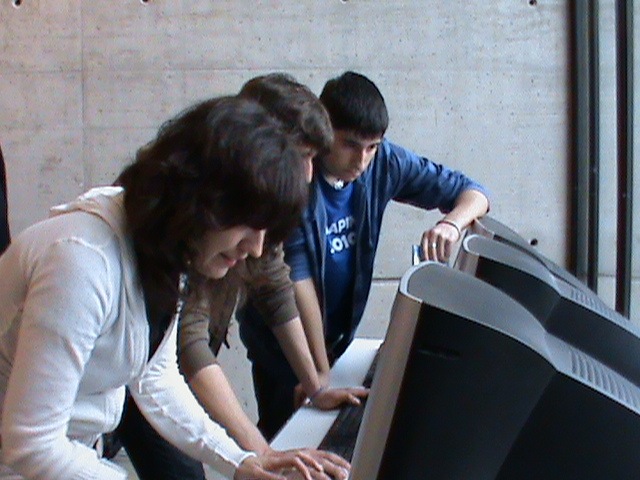} &
\includegraphics[height=0.59in]{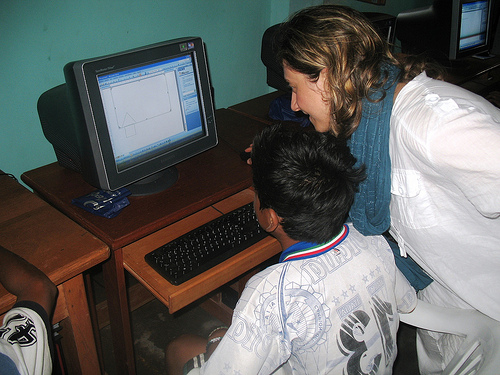} \\
\multicolumn{7}{c}{Balancing skin color:}\\
\includegraphics[height=0.59in]{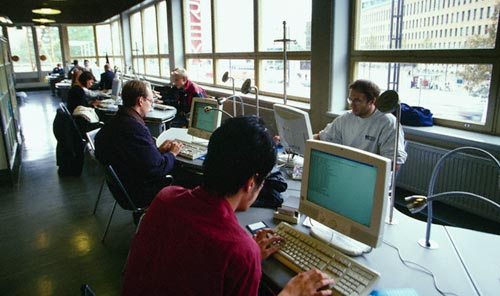} &
\includegraphics[height=0.59in]{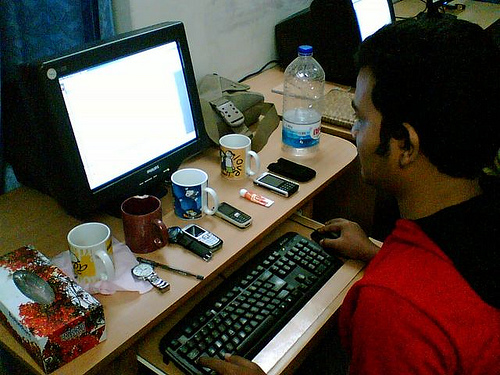} &
\includegraphics[height=0.59in]{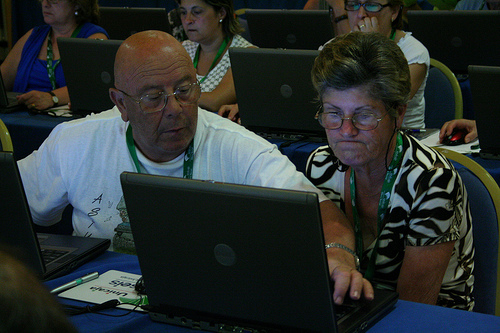} &
\includegraphics[height=0.59in]{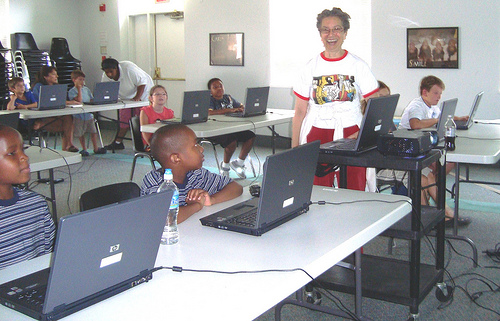} &
\includegraphics[height=0.59in]{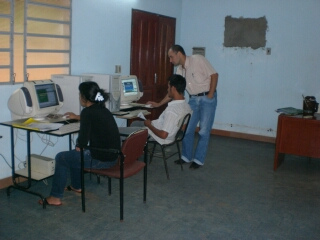} &
\includegraphics[height=0.59in]{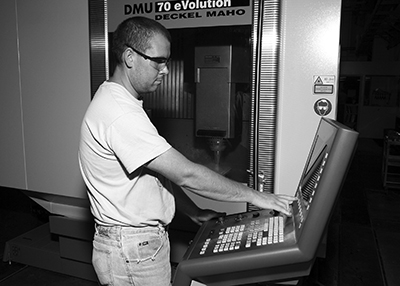} &
\includegraphics[height=0.59in]{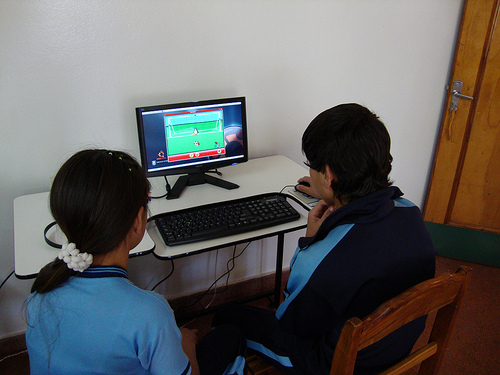} \\
\multicolumn{7}{c}{Balancing age:}\\
\includegraphics[height=0.59in]{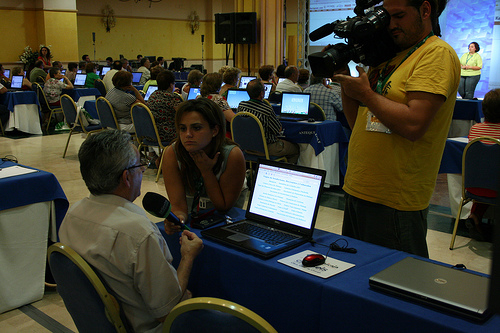} &
\includegraphics[height=0.59in]{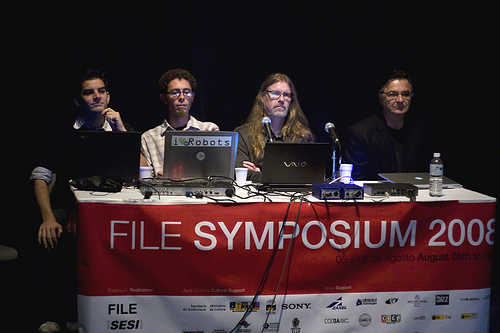} &
\includegraphics[height=0.59in]{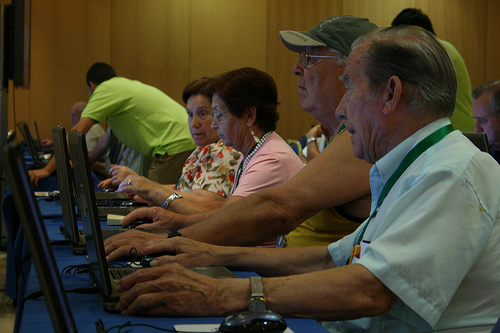} &
\includegraphics[height=0.59in]{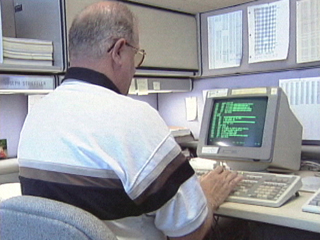} &
\includegraphics[height=0.59in]{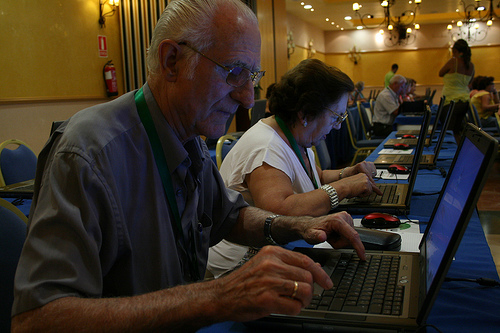} &
\includegraphics[height=0.59in]{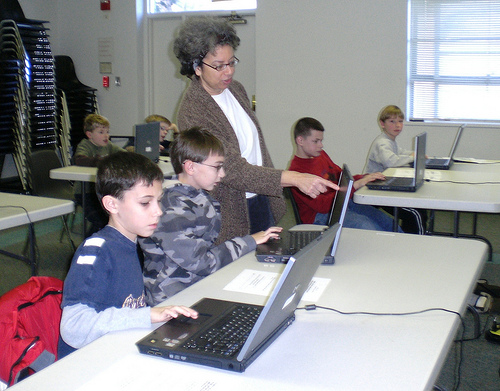} &
\includegraphics[height=0.59in]{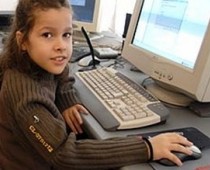} \\
    \end{tabular}
    \end{small}
    \caption{The distribution of images in the ImageNet synset \texttt{programmer} before and after balancing to a uniform distribution.}
    \label{fig:balancing}
\end{figure*}

\subsection{Results of the demographic analysis}
We annotated demographics on the 139 synsets that are considered both safe (Sec.~\ref{sec:synsets}) and imageable (Sec.~\ref{sec:imageability}) and that contain at least 100 images. 
We annotated 100 randomly sampled images from each synset, summing up to 13,900 images. Due to the presence of multiple people in an image, each image may have more than one category for each attribute. We ended up with 43,897 attribute categories annotated (14,471 annotations for gender; 14,828 annotations for skin; and 14,598 annotations for age). This was the result of obtaining and consolidating 109,545 worker judgments.

Fig.~\ref{fig:attrs_results} shows the distribution of categories for different synsets, which mirrors real-world biases. 
For gender, there are both male-dominated synsets and female-dominated synsets; but the overall pattern across all synsets reveals underrepresentation of female, as the blue area in Fig.~\ref{fig:attrs_results} (Left) is significantly larger than the green area. Relatively few images are annotated with the \textit{Unsure} category except a few interesting outliers: \texttt{birth} (58.2\% images labeled \textit{Unsure}) and \texttt{scuba diver} (16.5\%). The gender cues in these synsets obscured because \texttt{birth} contain images of newborn babies, and \texttt{scuba diver} contains people wearing diving suits and helmets.

The figure for skin color (Fig.~\ref{fig:attrs_results} Middle) also presents a biased distribution, highlighting the underrepresentation of people with dark skin. The average percentage of the \textit{Dark} category across all synsets is only 6.2\%, and the synsets with significant portion of \textit{Dark} align with stereotypes: \texttt{rapper} (66.4\% images labeled \textit{Dark}) and \texttt{basketball player} (34.5\%). An exception is \texttt{first lady} (51.9\%), as most images in this synset are photos of Michelle Obama, the First Lady of the United States when ImageNet was being constructed.

\subsection{Limitations of demographic annotation}

Given the demographic analysis, it is desired to have a constructive solution to improve the diversity in ImageNet images. Publicly releasing the collected attribute annotations would be a natural next step. This would allow the research community to train and benchmark machine learning algorithms on different demographic subsets of ImageNet, furthering the work on machine fairness.

However, we have to consider that the potential mistakes in demographics annotations are harmful not just for the downstream visual recognition models (as all annotation mistakes are) but to the people depicted in the photos. Mis-annotating gender, skin color, or age can all cause significant distress to the photographed subject. Gender identity and gender expression may not be aligned (similarly for skin color or age), and thus some annotations may be incorrect despite our best quality control efforts. So releasing the image-level annotations  may not be appropriate in this context. 

\subsection{Methodology for increasing image diversity}
\label{sec:balancing_methodology}

We aim for an alternative constructive solution, one that strikes a balance between advancing the community's efforts and preventing additional harm to the people in the photos. One option we considered is internally using the demographics for targeted data collection, where we would find and annotate additional images to re-balance each synset. However, with the known issues of bias in search engine results~\cite{noble2018algorithms} and the care already taken by the ImageNet team to diversify the images for each synset (Sec.~\ref{sec:pipeline}), this may not be the most fruitful route.

Instead, we propose to release a Web interface that automatically re-balances the image distribution within each synset, aiming for a target distribution of a single attribute (e.g., gender) by removing the images of the overrepresented categories. There are two questions to consider: first, what is an appropriate target distribution, and second, what are the privacy implications of such balancing. 

First, identifying the appropriate target distribution is challenging and we leave that to the end users of the database. For example, for some applications it might make sense to produce a uniform gender distribution, for example, if the goal is to train an activity recognition model with approximately equal error rates across genders. In other cases, the goal might be to re-balance the data to better mimic the real-world distribution of gender, race or age in the category (as recorded by census data for example) instead of using the distribution exaggerated by search engines. 
Note that any type of balancing is only feasible on synsets with sufficient representation within each attribute category. For example, the synset \texttt{baby} naturally does not contain a balanced age distribution. Thus, we allow the user to request a subset of the categories to be balanced; for example, the user can impose equal representation of the three adult categories while eliminating the \textit{Child} category. 

Second, with regard to privacy, there is a concern that the user may be able to use this interface to infer the demographics of the \emph{removed} images. For example, it would be possible to visually analyze a synset, note that the majority of people within the synset appear to be female, and thus infer that any image removed during the gender-balancing process are annotated as female. To mitigate this concern, we always only include 90\% of images from the minority category in the balanced images and discard the other 10\%. Further, we only return a balanced distribution of images if at least 2 attribute categories are requested (e.g.,  the user cannot request a female-only gender distribution) and if there are at least 10 images within each requested category. 

While we only balance the distribution of a single attribute (e.g., gender), it is desirable to balance across multiple attributes. However, it will result in too few images per synset after re-balancing. For example, if we attempt to balance both skin color and gender, we will end up with very few images. This creates potential privacy concerns with regard to being able to infer the demographic information of the people in the individual photos.

\subsection{Results and estimated impact of the demographic balancing on ImageNet}

Fig.~\ref{fig:balancing} provides one example of the effect of our proposed demographic balancing procedure on the synset \texttt{programmer}. Based on our analysis and statistics so far, and under the restrictions described in Sec.~\ref{sec:balancing_methodology}, we could offer such a balancing on 131 synsets for gender (ignoring the highly skewed \textit{Unsure} category and posing uniform distribution among \textit{Male} and \textit{Female}), 117 synsets for skin color (uniform distribution for the three categories), and 81 synsets for age (removing the \textit{Child} category and posing a uniform distribution for the other three age categories). Users can create customized balancing results for each synset by choosing the attribute categories to balance on.

\subsection{Limitations of the balancing solution}

The downside of this solution is that balancing the dataset instead of releasing the image-level attribute annotations makes it impossible to evaluate the error rates of machine learning algorithms on demographic subsets of the data, as is common in the literature~\cite{zhang2018mitigating,zhao2017men,hendricks2018women,buolamwini2018gender}. Nevertheless, this strategy is a better alternative than using the existing ImageNet \texttt{person} subtree (strong bias), releasing the image-level annotations (ethically problematic), or collecting additional images (technically impractical).

\section{Discussion}
We took the first steps towards filtering and balancing the distribution of the \texttt{person} subtree in the ImageNet hierarchy. The task was daunting, as with each further step of annotation and exploration, we discovered deeper issues that remain unaddressed. However, we feel that this is a significant leap forward from the current state of ImageNet. We demonstrate that at most 158 out of the 2,832 existing synsets should remain in the \texttt{person} subtree, as others are inappropriate categories for visual recognition and should be filtered out. Of the remaining synsets, 139 have sufficient data (at least 100 images) to warrant further exploration. On those, we provide a detailed analysis of the gender, skin color and age distribution of the corresponding images, and recommend procedures for better balancing this distribution. 

While 139 categories may seem small in comparison to the current set, it is nevertheless sufficiently large-scale to remain interesting to the computer vision community: e.g., the PASCAL dataset has only 20 classes~\cite{everingham2010pascal}, CelebA has 40 attributes~\cite{liu2015faceattributes}, COCO has 80 object categories~\cite{lin2014microsoft}, the fine-grained CUB-200 dataset has 200 bird species~\cite{WelinderEtal2010}. Further, note that the most commonly used subset of ImageNet is the set of 1,000 categories in the ImageNet Large Scale Visual Recognition Challenge (ILSVRC)~\cite{russakovsky2015imagenet}, which remains unaffected by our filtering: the three ILSVRC synsets of the \texttt{person} subtree are \texttt{bridegroom} (n10148035; safe, imageability 5.0), \texttt{ballplayer} (n09835506; safe, imageability 4.6) and \texttt{scuba diver} (n09835506; safe, imageability 5.0). 

There is still much remaining to be done outside the \texttt{person} subtree, as incidental people occur in photographs in other ImageNet synsets as well, e.g., in synsets of pets, household objects, or sports. It is likely that the density and scope of the problem is smaller in other subtrees than within this one, so the filtering process should be simpler and more efficient. We are releasing our annotation interfaces to allow the community to continue this work.

\begin{acks}
Thank you to Arvind Narayanan and Timnit Gebru for thoughtful discussions, to the graduate students who annotated the unsafe synsets, and to the Amazon Mechanical Turk workers for the imageability and demographic annotations. This work is supported by the National Science Foundation under Grant No. 1763642. The project began in fall 2018, the ImageNet site was temporarily placed under maintenance in January 2019, and the work was submitted for peer review in August 2019. This research was initiated, conceptualized, and executed solely by the authors.
\end{acks}

\clearpage
\setcounter{table}{0}
\renewcommand{\thetable}{\Alph{table}}
\setcounter{figure}{0}
\renewcommand{\thefigure}{\Alph{figure}}
\appendix

\section{Appendix}

    We include the annotation interfaces and additional results as promised in the main paper. We organize the the appendix according to the sections of the main paper for ease of reference.

\subsection{PROBLEM 1: STAGNANT CONCEPT VOCABULARY}
\label{appendix:1}

\begin{figure*}[t]
    \centering
    \includegraphics[width=2.0\columnwidth]{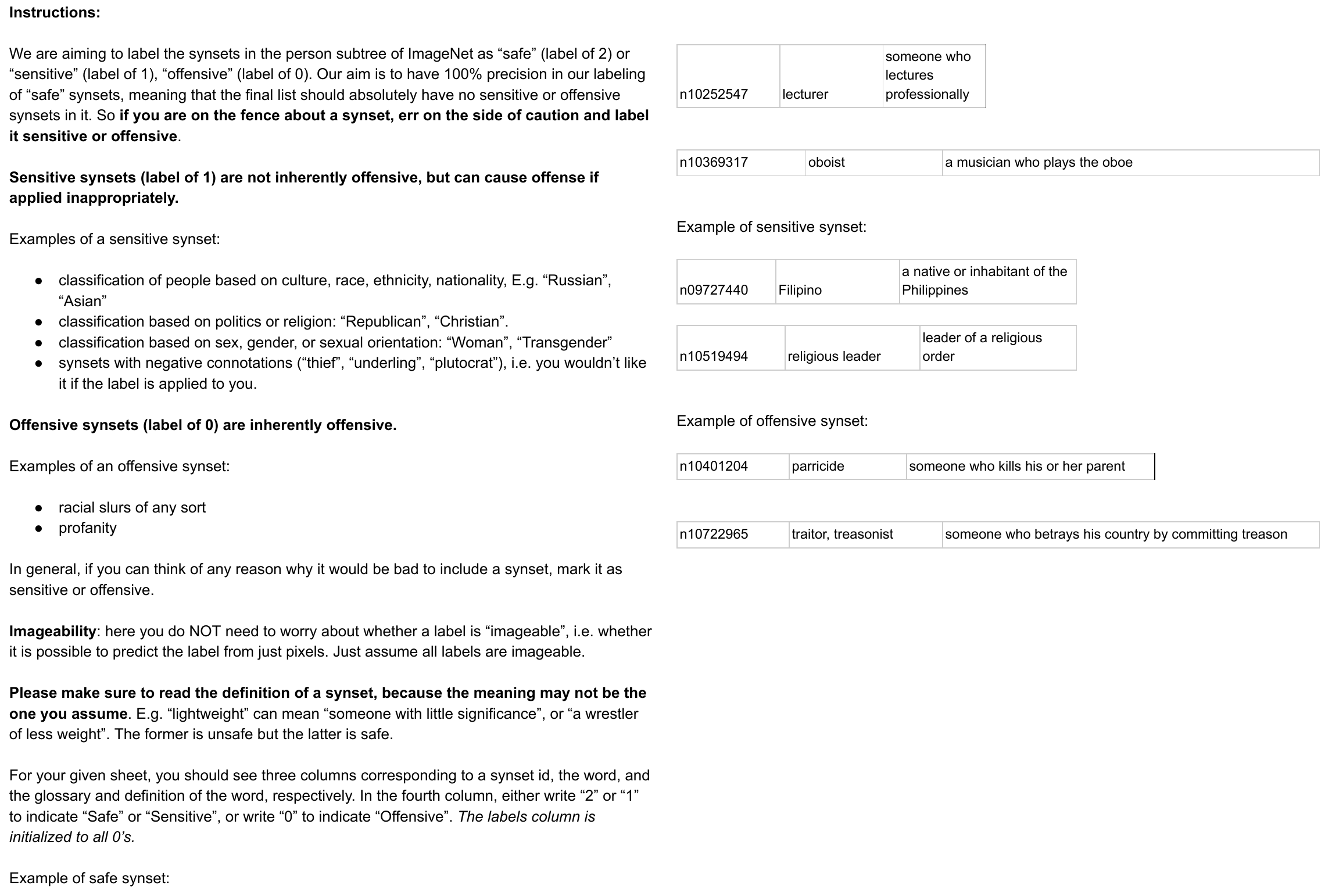}
    \caption{The instructions for annotating the offensiveness of synsets. The annotation was done in-house rather than using crowdsourcing, thus the user interface is kept simple.}
    \label{fig:offensive_ui}
\end{figure*}

The instructions used in-house to annotate the offensiveness of synsets are shown in Fig.~\ref{fig:offensive_ui}. We attach the synset IDs of the ``unsafe'' and ``safe'' synsets we have annotated. As before, we avoid explicitly naming the synsets, but the conversion from synset IDs to names can be found at \url{wordnet.princeton.edu/documentation/wndb5wn}.

\smallsec{Offensive synsets (1,593 in total)} 

\url{image-net.org/filtering-and-balancing/unsafe_synsets.txt}

\smallsec{Safe synsets (1,239 in total)} 

\url{image-net.org/filtering-and-balancing/safe_synsets.txt}

\subsection{PROBLEM 2: NON-VISUAL CONCEPTS}
\label{appendix:2}

\begin{figure*}[ht]
  \centering
  \includegraphics[width=2.0\columnwidth]{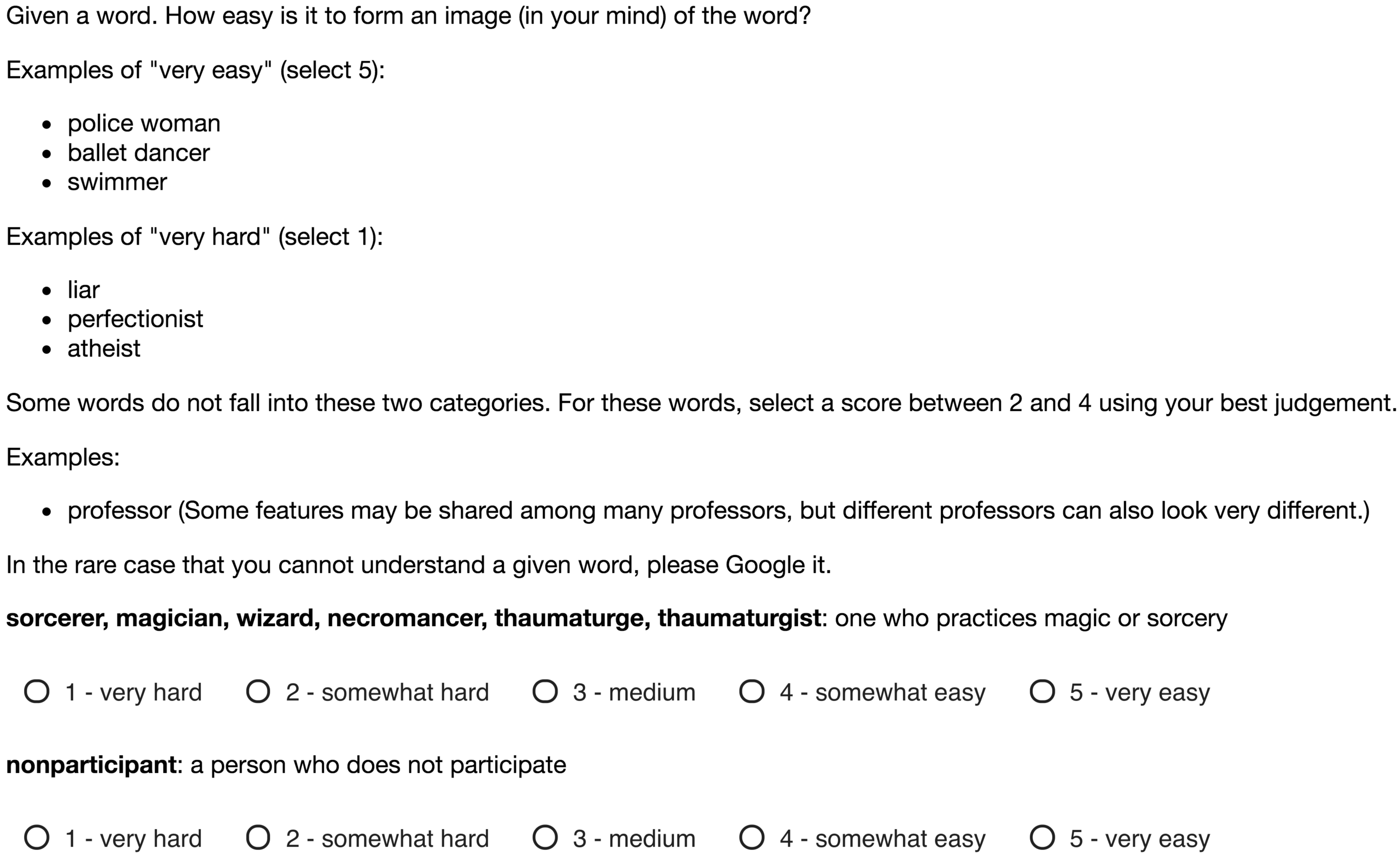} 
  \caption{User interface for crowdsourcing the imageability annotation.}
  \label{fig:imageability_ui}
  \Description{User interface for crowdsourcing the imageability annotation.}
\end{figure*}

\smallsec{Instructions} Fig.~\ref{fig:imageability_ui} shows the user interface for crowdsourcing imageability scores. 

\smallsec{Quality control} Table~\ref{table:gold_standard_imageability} lists the gold standard questions for quality control; half of them are obviously imageable (should be rated 5), and the other half are obviously non-imageable (should be rated 1). For a worker who answered a set of gold standard questions $Q$, we calculate the root mean square error of the worker as:
\begin{equation}
Error= \sqrt{\frac{1}{\vert Q \vert} \sum_{i \in Q}{(\widehat{x}_i - x_i)^2}}
\end{equation}
where $\widehat{x}_i$ is the rating from the worker and $x_i$ is the ground truth imageability for question $i$ ($\widehat{x}_i \in \{1,2,3,4,5\}, x_i \in \{1, 5\}$). If  $Error \geq 2.0$, we exclude all ratings of the worker.

Even after removing the answers from high-error workers, the raw ratings can still be noisy, which is partially attributed to the intrinsic subjectiveness in the imageability of synsets. We average multiple workers' ratings for each synset to compute a stable estimate of the imageability. However, it is tricky to determine the number of ratings to collect for a synset~\cite{sheng2008get}; more ratings lead to a more stable estimate but cost more. Further, the optimal number of ratings may be synset-dependent; more ambiguous synsets need a larger number of ratings. We devise a heuristic algorithm to determine the number of ratings dynamically for each synset.

Intuitively, the algorithm estimates a Gaussian distribution using the existing ratings, and terminates when three consecutive new ratings fall into a high-probability region of the Gaussian.  It automatically adapts to ambiguous synsets by collecting more ratings. 
Concretely, abusing notation from above (for simplicity of exposition), let $\widehat{\mathbf{x}} = [\widehat{x}_1, \widehat{x}_2, \widehat{x}_3, \dots, \widehat{x}_m]$ now be the sequence of ratings for a single synset from workers $1,2,3,\dots m$. After collecting $m \geq 4$ ratings, we partition the sequence into the last 3 ratings $\widehat{\mathbf{x}}_{new} = [\widehat{x}_{m-2}, \widehat{x}_{m-1}, \widehat{x}_m]$ and the rest $\widehat{\mathbf{x}}_{old} = [\widehat{x}_1, \widehat{x}_2, \dots, \widehat{x}_{m-3}]$. We compute the mean and standard deviation of $\widehat{\mathbf{x}}_{old}$ as  $\mu_{old}$ and $\sigma_{old}$, and we check whether the following holds:
\begin{equation}
   \forall x \in \widehat{\mathbf{x}}_{new}, \mu_{old} - \sigma_{old} \leq x \leq \mu_{old} + \sigma_{old}
\end{equation}
If it holds, the imageability annotations are approximately converging and we compute the final imageability score as the average of all ratings. Otherwise we collect more ratings and check again.

\begin{figure}[h]
\centering
\includegraphics[width=1.0\columnwidth]{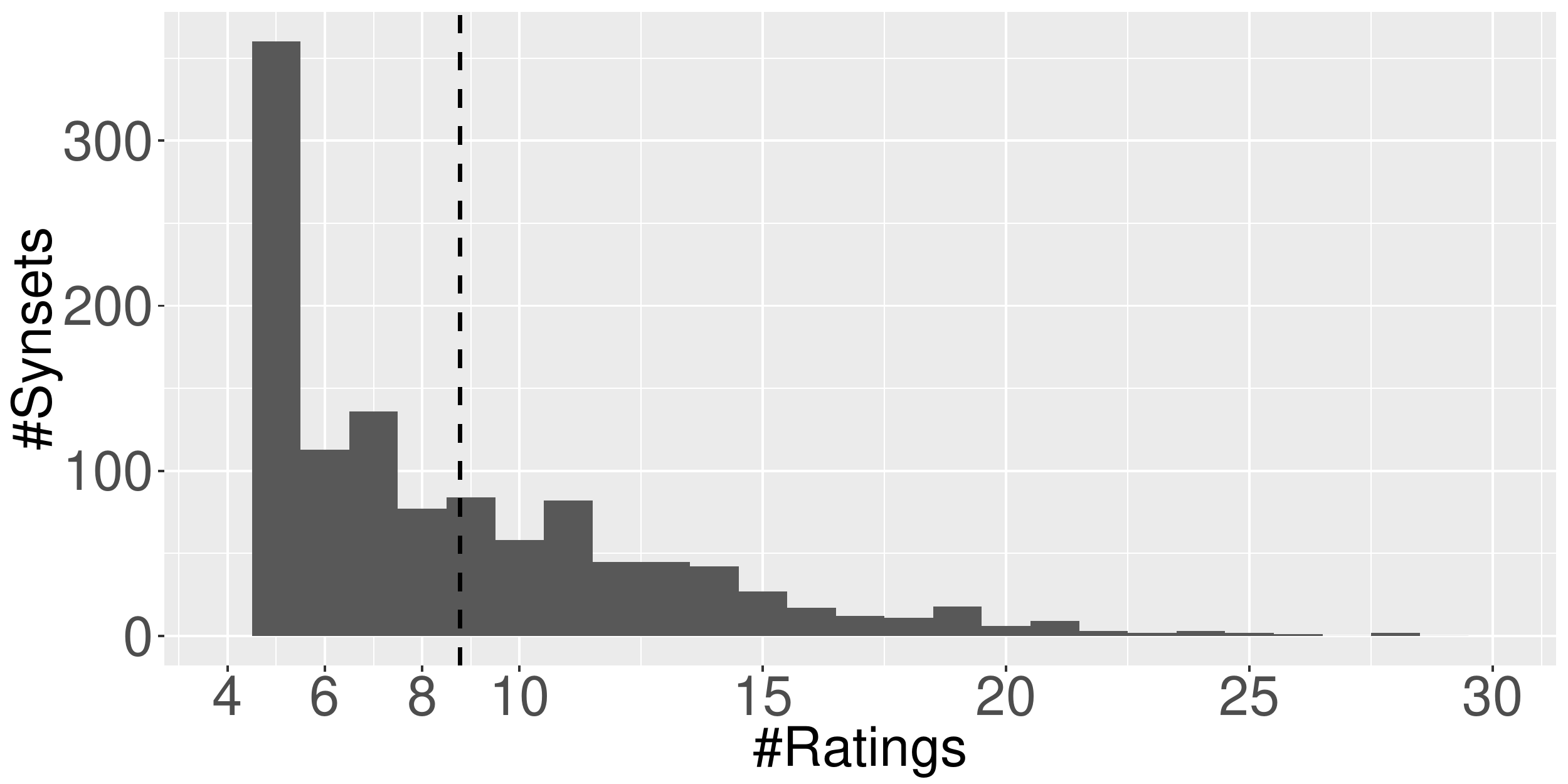}
\caption{The distribution of the number of raw imageability ratings collected for each synset. On average, the final imageability score of a synset is an average of 8.8 ratings.}
\label{fig:num_scores}
\end{figure}

Fig.~\ref{fig:num_scores} shows the number of ratings collected for the synsets. The average number is 8.8, and 72\% synsets need no more than 10 ratings. The file \url{image-net.org/filtering-and-balancing/imageability_scores.txt} includes the complete list of imageability scores for the 1,239 safe synsets in the \texttt{person} subtree.

\subsection{PROBLEM 3: LACK OF IMAGE DIVERSITY}
\label{appendix:3}

\begin{figure*}[t]
  \centering
  \includegraphics[width=2.0\columnwidth]{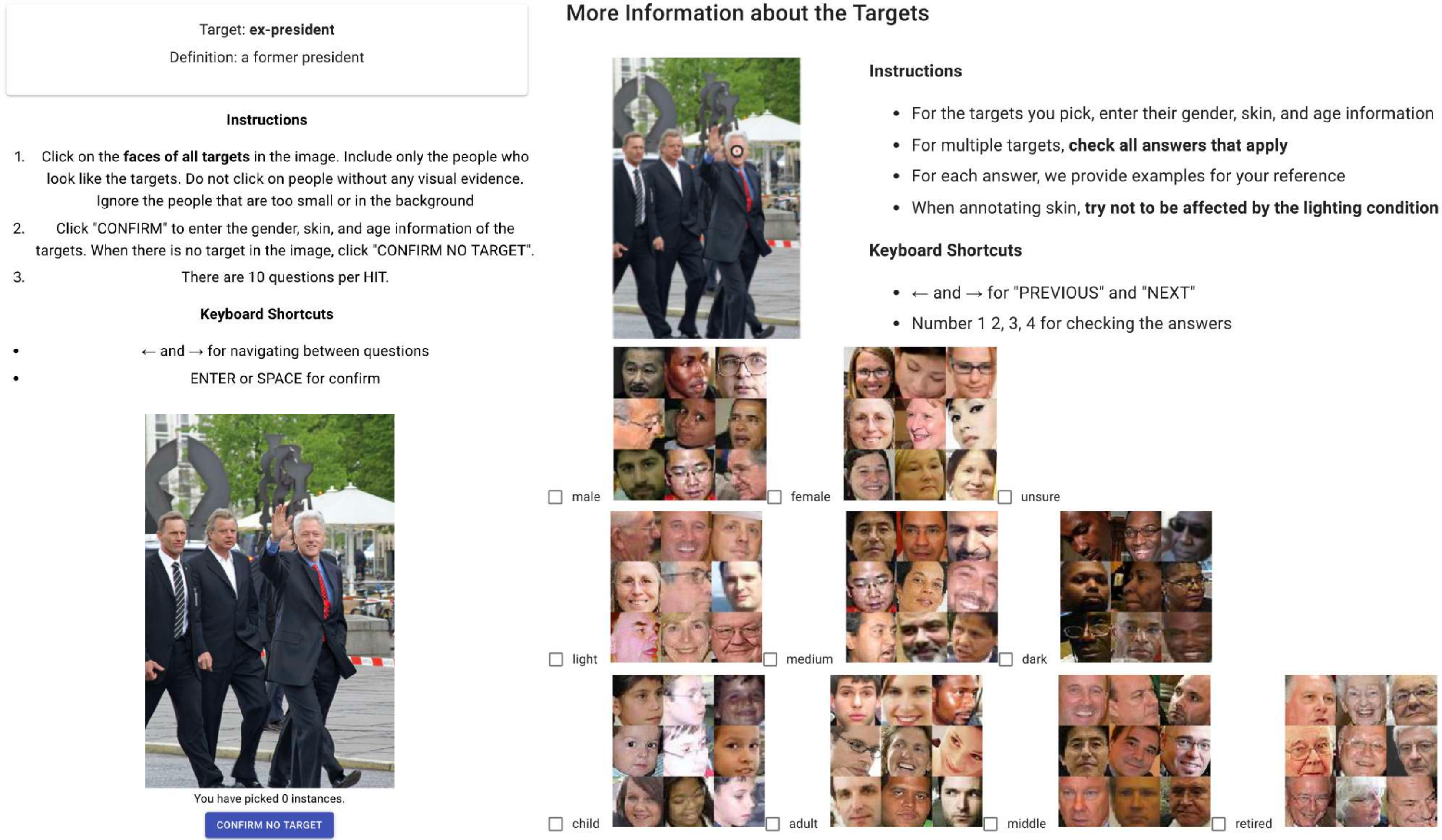} 
  \caption{User interface for crowdsourcing the demographics annotation.}
  \label{fig:demographics_ui}
  \Description{User interface for crowdsourcing the demographics annotation.}
\end{figure*}

The user interface used to annotate the protected attributes is shown in Fig.~\ref{fig:demographics_ui}.

\begin{table}[t]
\caption{Gold standard questions for quality control in imageability annotation.}
\label{table:gold_standard_imageability}
\begin{center}
\begin{small}
\resizebox{1.0\columnwidth}{!}{
\begin{tabular}{cccc}
\toprule
Synset ID & Synset & Ground truth imageability \\
\midrule
\texttt{n10101634} & football player, footballer &	5 \\
\texttt{n10605253} &	skier &	5 \\
\texttt{n09834885} &	ballet master &	5 \\
\texttt{n10366966} &	nurse  &	5 \\
\texttt{n10701644} &	tennis pro, professional tennis player  &	5 \\
\texttt{n09874725} &	bride &	5 \\
\texttt{n10772092} &	weatherman, weather forecaster &	5 \\
\texttt{n10536416} &	rock star  &	5 \\ 
\texttt{n09624168} &	male, male person & 5 \\
\texttt{n10087434} &	fighter pilot  &	5 \\
\texttt{n10217208} &	irreligionist  &	1 \\
\texttt{n10743356} &	Utopian &	1 \\
\texttt{n09848110} &	theist  &	1 \\
\texttt{n09755788} &	abecedarian  &	1 \\
\texttt{n09794668} &	animist  &	1 \\
\texttt{n09778927} &	agnostic &	1 \\
\texttt{n10355142} &	neutral  &	1 \\
\texttt{n10344774} &	namer  &	1 \\
\texttt{n09789898} &	analogist &	1 \\
\texttt{n10000787} &	delegate &	1 \\
\bottomrule
\end{tabular}
}
\end{small}
\end{center}
\end{table}

\bibliographystyle{ACM-Reference-Format}
\bibliography{acmart}

\end{document}